\documentclass[10pt,twocolumn,letterpaper]{article}

\usepackage[pagenumbers]{iccv} 

\definecolor{iccvblue}{rgb}{0.21,0.49,0.74}
\usepackage[pagebackref,breaklinks,colorlinks,allcolors=iccvblue]{hyperref}

\newcommand{\edit}[2]{\textcolor{red}{\st{#1}} \textcolor{blue}{#2}}

\usepackage{soul}
\usepackage{xcolor}
\usepackage{makecell}
\usepackage{multirow}

\usepackage{booktabs}
\usepackage{bm}
\usepackage{colortbl}
\usepackage{amssymb}
\usepackage{amsmath}

\def\paperID{1964} 
\def\confName{ICCV}
\def\confYear{2025}

\title{HOLa: Zero-Shot HOI Detection with Low-Rank Decomposed \\ VLM Feature Adaptation}

\author{%
  {Qinqian Lei$^1$}\quad
  {Bo Wang$^2$}  \quad
      {Robby T. Tan$^{1,3}$}  \\
  $^1$National University of Singapore \;
   $^2$University of Mississippi  \\
   $^3$ASUS Intelligent Cloud Services (AICS)\\
  \tt\small {qinqian.lei@u.nus.edu}  \; \tt\small {hawk.rsrch@gmail.com}  \; \tt\small {robby\_tan@asus.com} 
}

\begin{document}
\maketitle

\begin{abstract}
Zero-shot human-object interaction (HOI) detection remains a challenging task, particularly in generalizing to unseen actions. 
Existing methods address this challenge by tapping Vision-Language Models (VLMs) to access knowledge beyond the training data. 
However, they either struggle to distinguish actions involving the same object or demonstrate limited generalization to unseen classes.
In this paper, we introduce HOLa (Zero-Shot HOI Detection with Low-Rank Decomposed VLM Feature Adaptation), a novel approach that both enhances generalization to unseen classes and improves action distinction.
In training, HOLa decomposes VLM text features for given HOI classes via low-rank factorization, producing class-shared basis features and adaptable weights. 
These features and weights form a compact HOI representation that preserves shared information across classes, enhancing generalization to unseen classes. 
Subsequently, we refine action distinction by adapting weights for each HOI class and introducing human-object tokens to enrich visual interaction representations.
To further distinguish unseen actions, we guide the weight adaptation with LLM-derived action regularization.
Experimental results show that our method sets a new state-of-the-art across zero-shot HOI settings on HICO-DET, achieving an unseen-class mAP of 27.91 in the unseen-verb setting. 
Our code is available at \hyperlink{}{https://github.com/ChelsieLei/HOLa}. 
\end{abstract}

\section{Introduction}
\label{sec:intro}
\begin{figure}[t]
	\centering
	\includegraphics[width=0.98\linewidth]{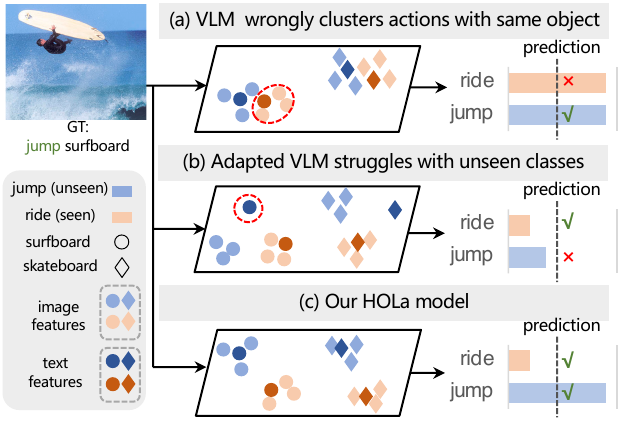}
	\caption{
		(a) HOI detection methods utilizing a frozen VLM feature space, struggle in distinguishing actions involving the same object~\cite{ning2023hoiclip, cao2024detecting, mao2024clip4hoi, Liao_2022_CVPR}. 
		(b) Approaches that adapt VLMs to HOI tasks based on supervision from seen HOI classes, struggle to generalize effectively to unseen classes~\cite{lei2024exploring, lei2024efficient}. 
        (c) Our HOLa model decomposes VLM features into class-shared basis and adaptable weights, improving generalization to unseen HOI classes and enhancing action distinction.
	}
	
	\label{fig: teaser}
\end{figure}

Human-object interaction (HOI) detection aims to identify humans and objects in images and predict their interactions, essential for many  applications~\cite{zhang2020perceiving,suhail2021energy, kong2022human, xie2022chore, lei2024few, cheng2023bottom, Zhang_2025_CVPR}. However, existing HOI detection struggles to generalize beyond seen classes, making zero-shot HOI detection a critical challenge~\cite{Liao_2022_CVPR, ning2023hoiclip, hou2020visual}.
Zero-shot HOI detection includes three primary settings: unseen composition, unseen object, and unseen action. 
In unseen composition HOI detection, actions and objects seen individually during training appear in novel combinations during testing, requiring models to recognize interactions beyond their training experience.
This setting can be addressed via compositional learning strategies that independently classify objects and actions~\cite{hou2020visual, hou2021detecting, hou2021affordance, lei2023efficient}. 
Unseen object HOI detection focuses on detecting interactions involving objects not seen during training, which can be tackled with open-vocabulary object detectors~\cite{minderer2022simple, gu2021open, zareian2021open}.
However, since these approaches assume that all actions are observed during training, they cannot generalize to unseen actions.

A promising direction is leveraging Vision-Language Models (VLMs) to harness their broad knowledge beyond HOI training data.
Existing methods following this approach fall into two main categories. The first maps HOI image features into the feature space of frozen VLMs~\cite{Liao_2022_CVPR, ning2023hoiclip, mao2024clip4hoi, cao2024detecting, li2024neural}, such as CLIP~\cite{radford2021learning}. 
However, since VLMs are primarily trained on large-scale datasets focused on object recognition and description~\cite{radford2021learning}, they struggle to differentiate actions, especially those involving the same object. 
This limitation is critical in zero-shot HOI detection, where methods~\cite{Liao_2022_CVPR, ning2023hoiclip, mao2024clip4hoi, cao2024detecting, li2024neural} depend on frozen VLMs to generalize to unseen classes without direct training supervision. 
Without explicitly refining VLM features to emphasize action-specific details, frozen VLMs lack the ability to distinguish actions involving the same object, as their representations primarily capture object semantics rather than nuanced action differences.

The second category~\cite{lei2024exploring, lei2024efficient} adapts VLMs for HOI detection using prompt learning~\cite{khattak2023maple, zang2022unified, shu2022test, jia2022visual, zhou2022learning}, reducing trainable parameters and computational costs. 
However, since training data provides annotations only for seen classes, these methods 
either do not explicitly leverage learned features for unseen classes~\cite{lei2024exploring} or lack sufficient supervision to guide generalization~\cite{lei2024efficient}. 
As a result, their adaptations fail to effectively generalize to unseen classes.
Moreover, they adapt the VLM visual encoder without explicitly modeling interactions or rely on limited interaction patterns, which limits the effectiveness of action distinction.

To improve generalization to unseen classes while enhancing action distinction, we propose HOLa (Zero-Shot HOI Detection with Low-Rank Decomposed VLM Feature Adaptation), which introduces two key ideas: (1) improved generalization via low-rank decomposition of VLM features to capture class-shared information, and (2) enhanced action distinction through adapted decomposed VLM features with LLM-derived action regularization and human-object tokens that enrich interaction representations.

We first apply low-rank factorization to VLM text features derived from HOI class descriptions, decomposing them into basis features and adaptable weights.
The basis features capture class-shared information, forming a foundation for HOI representation, while adaptable weights capture information unique to each HOI class. 
By combining basis features with adaptable weights, we construct a compact representation that retains shared knowledge across HOI classes, improving generalization to unseen classes.

However, distinguishing actions involving the same object remains challenging. The adaptable weights are derived from VLM text features, which struggle to encode fine-grained action differences.
To address this, we refine these weights through an adaptation process that enhances action distinction.
However, since training data only supervise seen classes, direct weight adaptation struggles to distinguish unseen actions.
To mitigate this, we introduce LLM-derived action regularization, which constrains the weight adaptation process to enhance generalization beyond seen actions.
Additionally, we integrate human-object tokens to encode spatial and appearance cues, further enriching visual interaction representations for action differentiation.

Fig.~\ref{fig: teaser} illustrates the key differences between our approach and existing HOI detection methods.
Fig.~\ref{fig: teaser} (a) shows how existing methods rely on a frozen VLM feature space, clustering actions involving the same object together, which leads to ambiguities~\cite{ning2023hoiclip, cao2024detecting, mao2024clip4hoi, Liao_2022_CVPR}.
Fig.~\ref{fig: teaser} (b) depicts methods that adapt VLMs to HOI settings, where direct adaptation struggles to generalize to unseen classes effectively, as training data provides ground truths only for seen classes~\cite{lei2024exploring, lei2024efficient}.
%
In contrast, our method enhances action distinction and improves generalization to unseen classes through novel decomposed VLM feature adaptation.
In summary, our contributions are:
\begin{itemize}
\item HOLa, a novel low-rank decomposed VLM feature adaptation method for zero-shot HOI detection, enhancing generalization to unseen classes, including unseen actions, while improving action distinction.
\item A novel low-rank factorization approach to decompose and reconstruct VLM text features, generating a compact representation that captures class-shared information and improves generalization.
\item Weight adaptation with LLM-derived action regularization and human-object tokens to refine interaction representation, both strengthening action distinction.
\end{itemize}
Extensive experiments on various zero-shot HOI settings demonstrate that our method achieves new state-of-the-art performance, reaching an unseen-class mAP of 27.91 in the unseen-verb setting on the HICO-DET dataset.

\begin{figure*}[t]
	\centering
	\includegraphics[width=0.98\textwidth]{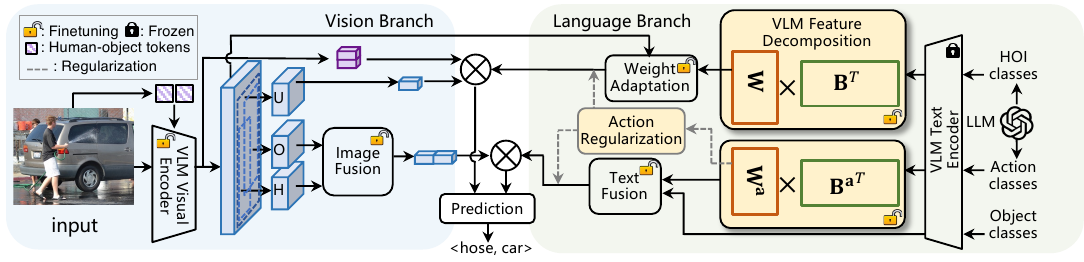}
	\caption{
		Overview of our HOLa. 
        In the language branch, VLM text HOI features are decomposed into HOI basis features $\bm{\mathrm{B}}$ and HOI weights $\bm{\mathrm{W}}$. Similarly, action features are decomposed into action basis features $\bm{\mathrm{B}^a}$ and action weights $\bm{\mathrm{W}^a}$, where $\bm{\mathrm{B}^a}$ is selected from $\bm{\mathrm{B}}$. 
        The weight adaptation updates $\bm{\mathrm{W}}$ with LLM-derived action regularization $\bm{\mathrm{W}^a}$, containing LLM-generated action information. The text fusion module combines action and object text features. 
		In the vision branch, we adapt the VLM visual encoder with human-object tokens. Then we crop humans, objects, and HOI union regions from encoder output (``H, O, U'' in the figure). The image fusion module then combines human and object features. 
        The prediction combines vision and language branches. 
	} \label{fig: overview}
\end{figure*}

\section{Related Work}
\label{sec:2_related_work}
\noindent \textbf{Human-Object Interaction Detection}
HOI detection methods are broadly categorized into one-stage and two-stage approaches.
One-stage methods predict all outputs simultaneously, including human and object bounding boxes, object categories, and interaction classes. Recent advancements leverage transformer architectures, achieving strong performance~\cite{chen2021qahoi, qu2022distillation, tamura2021qpic, zou2021end, yuan2022rlip, xie2023category, kim2023relational, luo2024discovering}.
In contrast, two-stage methods split HOI detection into object detection and HOI classification~\cite{gkioxari2018detecting, hou2020visual, zhang2021spatially, lei2023efficient, wu2024exploring}. This separation allows each module to specialize, leading to a more efficient process~\cite{zhang2022efficient}. Recent works have also integrated transformer architectures into two-stage designs, demonstrating promising results~\cite{zhang2023exploring, park2023viplo, wang2024bilateral}.

\vspace{0.2cm}
\noindent \textbf{Zero-Shot HOI Detection}
Prior efforts in zero-shot HOI detection mainly address cases where action and object classes are seen individually but not in combination, using compositional learning strategies~\cite{hou2020visual, hou2021detecting, hou2021affordance}. However, they struggle with unseen actions, as they require training representations for all actions and objects.
A promising alternative is leveraging VLMs to incorporate external knowledge, especially for unseen classes. 
Several methods align HOI visual features with frozen VLM text features~\cite{Liao_2022_CVPR, ning2023hoiclip, cao2024detecting, mao2024clip4hoi}, but VLM features, trained primarily for object recognition~\cite{radford2021learning}, often fail to capture actions details, resulting in HOI classes involving the same object being clustered together despite differing actions.
While our method also uses the frozen VLM text encoder, it avoids this issue by decomposing the features into class-shared basis and adaptable weights, enabling the model to refine action-specific differences and enhance action distinction.

\vspace{0.2cm}
\noindent \textbf{Vision-Language Adaptations for HOI detection}
VLMs are widely used in tasks like image classification, semantic segmentation, etc, due to their strong image understanding capabilities (e.g.~\cite{alayrac2022flamingo, li2022blip, li2023blip, radford2021learning, liu2023visual, liu2023improved}).
Various adaptation techniques have been explored, including prompt tuning~\cite{zhou2022conditional, zhou2022learning, jia2022visual, shu2022test, zang2022unified, khattak2023maple} and learnable adapters~\cite{zhang2021tip, gao2024clip, sung2022vl, yang2024mma, hu2021lora}.
In HOI detection, recent methods~\cite{lei2023efficient, lei2024exploring, lei2024efficient} adapt VLMs efficiently but struggle to generalize to unseen classes. Specifically, ADA-CM~\cite{lei2023efficient} performs poorly in the unseen-verb setting due to its memory-based design, which depends on ground-truth training data for each action, limiting generalization to unseen actions. Similarly, EZ-HOI~\cite{lei2024efficient} and CMMP~\cite{lei2024exploring} fail to generalize beyond seen classes, as their learnable prompts are fine-tuned solely on seen-class training data.
This underscores the need for approaches that enhance generalization to unseen classes. Our method addresses this by introducing low-rank VLM feature decomposition and LLM-guided action regularization, avoiding reliance on memory-based adaptation (as in ADA-CM)  and addressing the lack of generalization mechanism or supervision (as in CMMP and EZ-HOI).

\section{Proposed Method}
\label{sec: method}

Our method follows the two-stage HOI detection framework, which uses an off-the-shelf object detector~\cite{carion2020end} and focuses primarily on interaction classification~\cite{lei2023efficient, hou2020visual, hou2021detecting, lei2024exploring, zhang2021spatially, zhang2022efficient}.
Let $\mathcal{A} = \{a_1, a_2, \cdots, a_{N_a}\}$ be the set of actions and $\mathcal{O} = \{o_1, o_2, \cdots, o_{N_o}\}$ be the set of objects.
$N$ is the total number of HOI classes.
The HOI set consists of all given action-object pairs $\mathcal{C} = \{{\rm hoi}_i = (a_{i_a}, o_{i_o}) \mid a_{i_a} \in \mathcal{A}, o_{i_o} \in \mathcal{O}\}$.

Zero-shot HOI detection requires HOI models to generalize to unseen classes during inference~\citep{Liao_2022_CVPR}.
Let $\mathcal{S}$ be the set of seen HOI classes, and let $\mathcal{U} = \{{\rm hoi}_i \mid {\rm hoi}_i \notin \mathcal{S}, {\rm hoi}_i \in \mathcal{C}\}$ denote the set of unseen HOI classes. Our method follows standard zero-shot HOI settings, where the HOI class names of both seen and unseen classes are available during training. However, ground truth annotations (i.e., object bounding boxes, object labels, and their interaction labels) are provided only for seen classes and not for unseen ones~\cite{hou2020visual, hou2021affordance, wu2023end, ning2023hoiclip}.

As shown in Fig.~\ref{fig: overview}, our method consists of a language branch and a vision branch. The language branch decomposes VLM text features from HOI and action class names to capture class-shared information, enhancing generalization. These features are then refined through weight adaptation and text fusion modules respectively, incorporating LLM-derived action regularization to enhance action distinction. The vision branch utilizes a VLM visual encoder with our proposed human-object tokens to enhance interaction representation. Finally, the adapted language and vision features are fused for HOI prediction.

\subsection{VLM Feature Decomposition and Adaptation}
\label{sec: feat_decomp}

\noindent \textbf{VLM Feature Decomposition }
Given HOI class names $\mathcal{C}$, an LLM~\cite{dubey2024llama} generates detailed descriptions for each class, tapping its extensive knowledge beyond simple names.
These descriptions are fed into the VLM text encoder, producing $\bm{\mathrm{F}} = [\bm{f_1}, \bm{f_2}, \dots, \bm{f_N}]^\top \in \mathcal{R}^{N \times d}$, where $d$ is the feature dimension. Each $\bm{f_i}$ corresponds to HOI class ${\rm hoi}_i = (a_{i_a}, o_{i_o})$ and carries richer, more nuanced HOI class information.

In feature decomposition, we aim to obtain basis features and corresponding weights.
Let the learnable weights be $\bm{\mathrm{W}} = [\bm{w_1}, \bm{w_2}, \dots, \bm{w_N}]^\top \in \mathcal{R}^{N \times m}$ and the learnable basis features be $\bm{\mathrm{B}} = [\bm{b_1}, \bm{b_2}, \dots, \bm{b_m}] \in \mathcal{R}^{d \times m}$.
The low-rank factorization of VLM text features $\bm{\mathrm{F}}$ is formulated as:
\begin{equation}
\label{eq: recon1}
	L_{\rm recon}^1 = \min_{ \bm{\mathrm{W}},  \bm{\mathrm{B}}  } \lVert \bm{\mathrm{F}} - \bm{\mathrm{W}}   \bm{\mathrm{B}}^\top \rVert_2.
\end{equation}
Each $\bm{w_i} \in \mathcal{R}^{m \times 1}$ in $\bm{\mathrm{W}}$ corresponds to a specific HOI class ${\rm hoi}_i$, while all $\bm{b_i} \in \mathcal{R}^{d \times 1}$ in $\bm{\mathrm{B}}$ are shared by all given HOI classes (i.e., both seen and unseen classes). 

We apply $L_1$ loss  to the weights \( \bm{\mathrm{W}} \) during low-rank factorization to encourage sparsity: 
\begin{equation}
	\label{eq: sparse_hoi}
	L_{\rm sparse}^1 = \min_{\bm{w_i}} \sum_{i=1}^{i=N} \lVert \bm{w_i} \rVert_1.
\end{equation}
We impose an orthogonal constraint on $\bm{\mathrm{B}}$ to enforce feature orthogonality, enabling it to capture disentangled features~\cite{rodriguez2016regularizing} shared across HOI classes:
\begin{equation}
\label{eq: ort1}
	L_{\rm ort}=  \min_{\bm{b_i}, \bm{b_j}} \sum_{i=1, i \neq j}^{i=m} \sum_{j=1}^{j=m} \bm{b_i}^\top \bm{b_j}. 
\end{equation}
We reconstruct VLM text features using $\bm{\mathrm{W}} \bm{\mathrm{B}}^\top$ in a compact representation, highlighting shared features across HOIs and improving generalization to unseen classes.

\vspace{0.3cm}
\noindent \textbf{Weight Adaptation }
Since VLM text features primarily focus on object information rather than actions~\cite{radford2021learning}, directly using the reconstructed features $\bm{\mathrm{W}} {\bm{\mathrm{B}}}^\top$ for HOI prediction is suboptimal, as it struggles to distinguish actions involving the same object.
To address this, we introduce a text adapter to refine $\bm{\mathrm{W}} {\bm{\mathrm{B}}}^\top$ specifically for HOI tasks. 

First, we introduce down projection layers at the text adapter input and up projection layers at its output to reduce feature dimensions for efficient adaptation.
The input $\bm{\mathrm{W}} \overline{\bm{\mathrm{B}}}^\top$, where $\overline{\bm{\mathrm{B}}}$ represents the fixed basis features without gradients, first passes through a self-attention module. 
It then enters a cross-attention mechanism, where the self-attention output serves as the query, and the key and value are derived from the input image features $f_{\rm img}$ to integrate visual context.
Finally, a residual connection adds the input $\bm{\mathrm{W}} \overline{\bm{\mathrm{B}}}^\top$ back to the output, refining the adaptation while preserving input information.
The adapted weights $\bm{\mathrm{W}}$, when multiplied by $\overline{\bm{\mathrm{B}}}^\top$, yield the reconstructed HOI text features, which we denote as the adapted features $\bm{\hat{\mathrm{F}}}$.

Keeping $\overline{\bm{\mathrm{B}}}$ unchanged is essential for preserving shared information across HOI classes from the VLM.
Thus, the text adapter fine-tunes only the weights $\bm{\mathrm{W}}$ for each HOI class, optimizing them with the action classification loss in Eq.~(\ref{eq: fl}) to enhance action distinction while maintaining shared HOI knowledge.

\begin{figure}[t]
	\centering
	\includegraphics[width=0.98\linewidth]{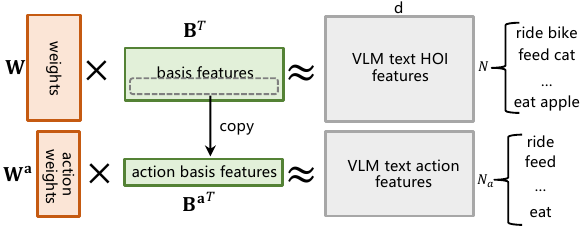}
	\caption{VLM text feature decomposition illustration. We first decompose HOI features into basis features $\bm{\mathrm{B}}$ and weights $\bm{\mathrm{W}}$. Then we obtain action-related basis features from $\bm{\mathrm{B}}$, which can be used to reconstruct the action features. 
	}
	
	\label{fig: feat_decomp}
\end{figure}

\subsection{Language Branch}

As shown in Fig.~\ref{fig: overview}, our language branch processes both HOI class names and separate action and object names. HOI text features, derived from HOI class names, capture richer interaction information, while weight adaptation enhances action distinction. However, its dependence on seen-class training data limits generalization.
To address this, we introduce LLM-derived action regularization to guide adaptation, improving unseen action distinction. We then fuse the action text features with object text features for a more robust representation.

\vspace{0.3cm}
\noindent \textbf{LLM-Derived Action Regularization }
Our idea is to leverage LLM-derived action regularization to constrain adaptable weights during adaptation, preventing an over-reliance on seen-class information.
First, we utilize an LLM to generate detailed action descriptions for each action class in $\mathcal{A}$. These descriptions are then encoded into action text features using a VLM text encoder, producing $\bm{\mathrm{F}^a}=[\bm{f_1^a}, \bm{f_2^a}, \cdots, \bm{f_{N_a}^a}]^\top \in \mathcal{R}^{N_a \times d}$. 
Next, $\bm{\mathrm{F}^a}$ is decomposed into weights $\bm{\mathrm{W}^a} \in \mathcal{R}^{N_a \times k}$ and basis features $\bm{\mathrm{B}^a} \in \mathcal{R}^{d \times k}$, with the factorization process formulated as:
\begin{equation}
\label{eq: recon2}
	L_{\rm recon}^2 = \min_{ \bm{\mathrm{W}^a},  \bm{\mathrm{B}^a}  } \lVert \bm{\mathrm{F}^a} - \bm{\mathrm{W}^a}   {\bm{\mathrm{B}^a}}^\top \rVert_2, 
\end{equation}
where $ k \ll \min(d, N_a)$.
The decomposition includes a sparsity constraint on $\bm{\mathrm{W}^a}$, expressed as:
$
	L_{\rm sparse}^2 = \min_{\bm{w_i^a}} \sum_{i=1}^{i=N_a} \lVert \bm{w_i^a} \rVert_1.
$

The action basis $\bm{\mathrm{B}^{a}}$ is randomly sampled from $\bm{\mathrm{B}}$ at initialization and shared by both HOI and action reconstruction, so the action reconstruction enforces action cues in $\bm{\mathrm{B}^{a}}$.
We denote $\bm{b}_{i}$ as the $i$-th row of the matrix $\mathbf{B}$ that also belongs to the subset $\mathbf{B}^{a}$, and define the index set ${\mathcal{I}} = \{\,i \mid \bm{b}_{i}\in\mathbf{B}^{a}\,\}$.
Since $\bm{\mathrm{W}^a}$ encodes nuanced action information derived from action descriptions for both seen and unseen classes, we use it to regularize a matching subset of the adapted HOI weights $\bm{{\mathrm{W}}}$. 
Concretely, we extract the subset $ {\mathbf{W}}_{\mathrm{ar}} = \{ \hat{\bm{w}}_{i}' \mid i \in {\mathcal{I}} \}$, where $\hat{\bm{w}}_{i}'$ denotes the $i$-th column of the adapted matrix $\bm{\mathrm{W}}$. 
Finally, we align $\bm{{\mathrm{W}}_{\rm ar}}$ with $\bm{\mathrm{W}^a}$ by minimizing their KL divergence:
\begin{equation}
\label{eq: reg1}
	\begin{aligned}
		L_{\rm act}^1 = \min D_{\text{KL}} [\: \bm{{\mathrm{W}}_{\rm ar}} \: \parallel \:  \bm{\mathrm{W}^a} \: ], 
	\end{aligned}
\end{equation}
where $D_{\text{KL}}$ represents the KL divergence. 

This approach encourages the adapted action-related subset $\bm{{\mathrm{W}}_{\rm ar}}$ to align with the weight distributions of $\bm{\mathrm{W}^a}$, integrating action knowledge from LLM-derived action regularization.
As a result, the adaptation of $\bm{{\mathrm{W}}_{\rm ar}}$ is no longer limited to seen-class training data but is also guided by $\bm{\mathrm{W}^a}$, which extracts generalizable action information from LLM-generated descriptions. This enhances action distinction, particularly for unseen classes.

\begin{figure}[t]
	\centering
	\includegraphics[width=0.48\textwidth]{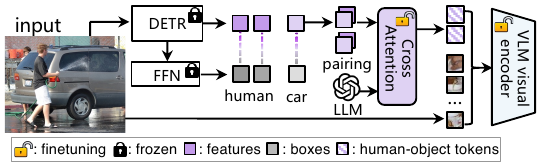}
	\caption{
		Human-object token illustration in the vision branch. 
	} \label{fig: ho_token}
\end{figure}

\vspace{0.3cm}
\noindent \textbf{Text Fusion }
While HOI text features capture overall human-object interactions, individual action and object text features provide finer details specific to each component. Fusing action and object text features combines these details, enriching the HOI representation with more precise object and action information.

Specifically, the input to the text fusion module is the concatenated action and object features:
$ {\rm concat}(\bm{\mathrm{W}^a}\,  {\bm{\mathrm{\overline{B}}^a}}^\top, \, \bm{{\mathrm{F}}^o})$, where the action feature is reconstructed from action basis features, ${\bm{\mathrm{\overline{B}}^a}}$, and action weights  $\bm{\mathrm{W}^a} $. 
The object features $\bm{\mathrm{F}^o}$ are obtained through VLM text encoders with a template input `` a photo of $\langle$object$\rangle$''. 
To reduce computational cost, we incorporate down and up projection layers.
The input features then pass through a self-attention module, integrating action and object information. Finally, a residual connection adds the input back to the output to retain input information.
We refer to the text fusion output as the adapted features $\bm{\hat{\mathrm{F}}^{\rm ao}}$.

We design adapted action weights $\bm{{\mathrm{W}}^a}$ to enhance action distinction while leveraging object information to enrich interaction understanding, capturing variations in actions across different object contexts (e.g., riding a bus vs. riding an elephant).
We apply LLM-derived action regularization during the adaptation of $\bm{{\mathrm{W}}^a}$ to improve unseen action distinction:
\begin{equation}
\label{eq: reg2}
	L_{\rm act}^2 = \min {D_{\text{KL}}} [\: \rm{adapted } \; \bm{{\mathrm{W}}^{a}} \: \parallel \:  \bm{{\mathrm{W}}^{a}}  \:].
\end{equation}

\subsection{Vision Branch} 
To enhance interaction representation and action distinction, we introduce human-object tokens in the vision branch, as shown in Fig.~\ref{fig: ho_token}.
 We utilize a pretrained DETR model~\cite{carion2020end} for object detection, identifying all detected humans ($h_i$) and objects ($o_j$), where $i$ and $j$ denote the indices of detected human and object instances, respectively.
 
For each detection, we extract human and object features from the DETR decoder, denoted as $f_{h_i}$ and $f_{o_j}$. We generate all possible human-object feature pairs, represented as $(f_{h_i}, f_{o_j})$.
We compute spatial features from paired human-object bounding boxes, denoted as $f_{{\rm ho}_{ij}}^{\rm spatial}$, which facilitate interaction recognition.
The human-object tokens are computed: $T_{{\rm ho}_{ij}} = \frac{f_{h_i} + f_{o_j}}{2} + f_{{\rm ho}_{ij}}^{\rm spatial}$,
which integrates both appearance and spatial cues to enhance interaction representation for a human-object feature pair $( f_{h_i}, f_{o_j})$. The human-object token set is ${T}_{\rm ho}$ where $  {T}_{\rm ho} = \{{T}_{{\rm ho}_{ij}} \mid 1 \leq i \leq n_h, 1\leq j \leq n_o  \}$.

After generating human-object tokens $T_{{\rm ho}_{ij}}$, we use an LLM to generate interaction prior knowledge features, 
which are combined with $T_{{\rm ho}_{ij}}$ as the output.
These human-object tokens, along with image patches, are then fed into the visual encoder, which outputs the adapted human-object tokens $ \hat{T}_{{\rm ho}_{ij}}$  and a global image feature map $f_{\rm img}^{\rm glb}$. 
We consequently obtain human, object and human-object union visual features, denoted as ${f}^h_{\rm img}, {f}^o_{\rm img}, {f}^u_{\rm img}$ from $f_{\rm img}^{\rm glb}$.
The image fusion module fuses human and object visual features, denoted as $f^{\rm ho}_{\rm img}$ enhancing interaction representation.
Additional details are provided in the supplementary.

\subsection{Training and Inference}
\label{subsec: train&inf}
\noindent \textbf{Training }
We can calculate the final action prediction $s_a$ using:
\begin{equation}
	\label{eq: score_act}
	\begin{aligned}
		s_a = \: & \gamma_1 * ( {\rm sim}({f}_{\rm img}^{u}, \bm{\hat{\mathrm{F}}} ) + {\rm sim}(\hat{T}_{\rm ho}, \bm{\hat{\mathrm{F}}} ) )* l_{u}^R + \\
		& \gamma_2 * {\rm sim}( f^{\rm ho}_{\rm img}, \bm{\hat{\mathrm{F}}^{\rm ao}}  ) * l_{ao}^R,  \\
	\end{aligned}
\end{equation}
where $l_{u}^R$ is the action class label corresponding to the adapted HOI features $\bm{\hat{\mathrm{F}}}$ and $l_{ao}^R$ is the action class label corresponding to $\bm{\hat{\mathrm{F}}^{\rm ao}}$. 
One action class can be associated with multiple HOI classes. If an action $a_{i_a}$ corresponds to $q$ HOI classes, the labels related to $a_{i_a}$ in $l_{u}^R$ and $l_{ao}^R$ are scaled by a normalization weight of $1/q$.
$\rm sim(\cdot \: | \: \cdot)$ indicates the cosine similarity. $\gamma_1, \gamma_2$ are hyper-parameters. 
We compute the classification loss $L_{\rm cls} $ for action classification:
\begin{equation}
\label{eq: fl}
	L_{\rm cls} = {\rm {\bm FL}}(s_a, s_{\rm GT}),
\end{equation}
where ${\rm FL}$ represents focal loss~\cite{lin2017focal} and $s_{\rm GT}$ represents the ground-truth action label.
We also apply a semantic loss $L_{\rm sem}$  to facilitate the action distinction (detailed in the supplementary).

Apart from supervision using HOI training data, we also include the following VLM feature decomposition constraints during training:
\begin{equation}
	\begin{aligned}
		\label{eq: fd_loss}
		L_{\rm fd}  & = \beta_1 (L_{\rm recon}^1 + L_{\rm recon}^2)  + \beta_2 (L_{\rm sparse}^1 + L_{\rm sparse}^2) \\
		& + \beta_3 L_{\rm ort} + \beta_4 (L_{\rm act}^1 + L_{\rm act}^2).
	\end{aligned}
\end{equation}
The final loss is obtained by:
$
	\label{eq: train_loss}
	L_{\rm total} = L_{\rm cls} + \alpha L_{\rm sem} +  L_{\rm fd}.
 $

\vspace{0.3cm}
\noindent \textbf{Generalization via Feature Decomposition}  
$L_{\mathrm{fd}}$ continuously update the basis $\bm{\mathrm{B}}$  in training by reconstructing the fixed VLM features and it serves as a regularizer to preserve unseen-class information in $\bm{\mathrm{B}}$.
Specifically, if $\bm{\mathrm{B}}$ is overly influenced by the classification loss $L_{\mathrm{cls}}$ and overfits to seen classes, its reconstruction error on the unseen classes of $\bm{\mathrm{F}}$ increases, and then $L_{\rm fd}$ penalizes it accordingly. 
In contrast, direct adaptation guided by $L_{\rm cls}$ alone loses unseen-class information in $\bm{\mathrm{F}}$ and overfits to seen classes. 
By decomposing $\bm{\mathrm{F}}$ into $\mathbf{W} \bm{\mathrm{B}}^{\top}$, each feature is reconstructed as a mixture of seen and unseen classes.
The class-shared basis $\bm{\mathrm{B}}$ 
enforces the adaptation to incorporate unseen information, thus preserving generalizable knowledge.

\vspace{0.3cm}
\noindent \textbf{Inference } ~At the test stage, we use an off-the-shelf detector~\cite{carion2020end}
for human and object detection and obtain human bounding boxes $b_h$, object bounding boxes $b_o$ and their related classification score $s_h$, $s_o$. From Eq.(~\ref{eq: score_act}), we can obtain the action classification score $s_a$. Finally, the HOI score for each human-object pair is computed as follows:
\begin{equation}
	s_{h,o}^a = (s_h * s_o)^\tau * \sigma(s_a),
\end{equation}
where $\sigma(\cdot)$ denotes sigmoid function, $\tau$ is a hyperparameter.

\section{Experiments}

\noindent  \textbf{Dataset }~
We evaluate our method on the HICO-DET dataset~\cite{chao2018learning}, a widely used benchmark for human-object interaction detection. Following prior works~\cite{Liao_2022_CVPR, ning2023hoiclip, mao2024clip4hoi, cao2024detecting}, we assess performance under four zero-shot HOI detection settings: unseen verb (UV), non-rare first unseen composition (NF-UC), rare first unseen composition (RF-UC), and unseen object (UO).
Additionally, we provide quantitative evaluations under the fully supervised setting on the HICO-DET and V-COCO datasets in the supplementary materials, where our method also achieves competitive results.

\begin{table}[t]
\centering
\resizebox{1\columnwidth}{!}{
	\begin{tabular}{l | c c c c }
		\toprule
		{Method} & HM & Unseen & Seen & Full \\
		\hline
        ADA-CM~\cite{lei2023efficient}~\scriptsize({ICCV'23}) & 21.67 & 17.33 & 28.92 & 27.29 \\
		GEN-VLKT~\cite{Liao_2022_CVPR}~\scriptsize({CVPR'22}) & 24.76 & 20.96 & 30.23 & 28.74 \\
		HOICLIP~\cite{ning2023hoiclip}~\scriptsize({CVPR'23}) & 27.69 & 24.30 & 32.19 & 31.09 \\
        UniHOI~\cite{cao2024detecting}~\scriptsize({NeurIPS'23}) & 26.62 & 22.18 & 33.29 & 30.87 \\
        LogicHOI~\cite{li2024neural}~\scriptsize({NeurIPS'23}) & 27.75 & 24.57 & 31.88 & 30.77 \\
        CLIP4HOI~\cite{mao2024clip4hoi}~\scriptsize({NeurIPS'23}) & 28.35 & 26.02 & 31.14 & 30.42 \\
        CMMP~\cite{lei2024exploring}~\scriptsize({ECCV'24}) & \underline{29.13} & \underline{26.23} & 32.75 & 31.84 \\
        EZ-HOI~\cite{lei2024efficient}~\scriptsize({NeurIPS'24}) & 28.69 & 25.10 & \underline{33.49} & \underline{32.32} \\
        \hline
    	\bm{$Ours$} (HOLa) & \textbf{31.09} & \textbf{27.91} & \textbf{35.09} & \textbf{34.09} \\
        \bottomrule
	\end{tabular}
}
\caption{
	Quantitative comparison of zero-shot HOI detection with state-of-the-art methods in the Unseen-Verb (UV) setting on HICO-DET. HM denotes the harmonic mean. 
}
\label{tab: UV}
\end{table}

\vspace{0.3cm}
\noindent  \textbf{Evaluation Metrics }~
Following established evaluation protocols for HOI detection, we evaluate our model using mean average precision (mAP)~\cite{lei2023efficient, ning2023hoiclip, Liao_2022_CVPR, zou2021end}.
To provide a more balanced assessment, we also report the harmonic mean (HM) metric~\cite{lei2024exploring}, which measures performance across both seen and unseen HOI classes, preventing the evaluation from being skewed by the larger number of seen classes.

\vspace{0.3cm}
\noindent  \textbf{Implementation Details }
For object detection, we use a pretrained DETR model~\cite{carion2020end} with a ResNet50~\cite{he2016deep} backbone, fine-tuned on HICO-DET, following existing zero-shot two-stage HOI detection methods~\cite{hou2020visual, bansal2020detecting, lei2023efficient}.
We set $m=71, k=35$ for feature decomposition, with a VLM feature dimension of $d=512$.
In Eq.(\ref{eq: score_act}), we use $\gamma_1 = 2.66, \gamma_2 = 2.66$.
In Eq.(\ref{eq: fd_loss}), we set $\beta_1=0.1, \beta_2=0.1, \beta_3=0.001, \beta_4=50$.
For training loss calculation $L_{\rm total}$, we assign $\alpha=80$.
The temperature $\tau$ in Eq.(\ref{eq: score_act}) is set to 1 during training and 2.8 during inference~\cite{zhang2021spatially, zhang2022efficient}.
To ensure a fair comparison with baseline methods leveraging VLMs, we use the CLIP model with the same ViT-B backbone~\cite{dosovitskiy2021an}. 
More details can be found in the supplementary.

\begin{table}[t]
\centering
\resizebox{1\columnwidth}{!}{
	\begin{tabular}{l | c c c c }
		\toprule
		{Method} & HM & Unseen & Seen & Full \\
		\hline
		GEN-VLKT~\cite{Liao_2022_CVPR}~\scriptsize({CVPR'22}) & 24.19 & 25.05 & 23.38 & 23.71 \\
		HOICLIP~\cite{ning2023hoiclip}~\scriptsize({CVPR'23}) & 27.22 & 26.39 & 28.10 & 27.75 \\
        ADA-CM~\cite{lei2023efficient}~\scriptsize({ICCV'23}) & 31.75 & 32.41 & 31.13 & 31.39 \\
        UniHOI~\cite{cao2024detecting}~\scriptsize({NeurIPS'23}) & 26.21 & 26.89 & 25.57 & 25.96 \\
        LogicHOI~\cite{li2024neural}~\scriptsize({NeurIPS'23}) & 27.34 & 26.84 & 27.86 & 27.95 \\
        CLIP4HOI~\cite{mao2024clip4hoi}~\scriptsize({NeurIPS'23}) & 29.77 & 31.44 & 28.26 & 28.90 \\
        CMMP~\cite{lei2024exploring}~\scriptsize({ECCV'24}) & 30.85 & 32.09 & 29.71 & 30.18 \\
        EZ-HOI~\cite{lei2024efficient}~\scriptsize({NeurIPS'24}) & \underline{32.03} & \underline{33.66} & \underline{30.55} & \underline{31.17} \\
        \hline
    	\bm{$Ours$} (HOLa) & \textbf{33.35} & \textbf{35.25} & \textbf{31.64} & \textbf{32.36} \\
		\bottomrule
	\end{tabular}
}
\caption{
	Quantitative comparison of zero-shot HOI detection with state-of-the-art methods in Non-Rare First Unseen-Composition (NF-UC) setting on HICO-DET. HM denotes the harmonic mean.}
\label{tab: NF-UC}
\end{table}

\subsection{Zero-Shot HOI Detection Evaluation}
\noindent \textbf{Unseen-Verb Setting }
Table~\ref{tab: UV} shows that our method outperforms all existing approaches, establishing a new state-of-the-art across every metric.
For unseen classes, our method achieves 27.91 mAP (+6.40\%) and an HM score of 31.09 (+6.73\%), surpassing the previous best result from CMMP~\cite{lei2024exploring}.
Our improved performance stems from our proposed modules, where low-rank factorization extracts class-shared basis features, allowing unseen classes to be effectively represented. Additionally, LLM-derived regularization mitigates reliance on training data supervision, leading to improved unseen action distinction.

\vspace{0.3cm}
\noindent \textbf{Unseen-Composition Setting }
The unseen composition setting includes two scenarios: non-rare first (NF-UC) and rare first (RF-UC).
In the NF-UC setting (Table~\ref{tab: NF-UC}), where common HOI classes are unseen, our method achieves the highest performance across all metrics, demonstrating its strong generalization ability. 
Our method surpasses the previous state-of-the-art by 1.59 mAP on unseen classes while achieving 31.64 mAP on seen classes.
In the RF-UC setting (Table~\ref{tab: RF-UC}), where rare HOI classes are unseen, our method outperforms all existing approaches, exceeding the previous best (CMMP~\cite{lei2024exploring}) by 1.59 mAP on unseen classes. Our method also achieves the best harmonic mean (HM) and full metric scores, ensuring a more balanced seen-unseen performance.
While CLIP4HOI~\cite{mao2024clip4hoi} scores 0.4 mAP higher on seen classes, our method demonstrates stronger generalization, outperforming it by 2.14 mAP on unseen classes.

\begin{table}[t]
\centering
\resizebox{1\columnwidth}{!}{
	\begin{tabular}{l | c c c c }
		\toprule
		{Method} & HM & Unseen & Seen & Full \\
		\hline
		GEN-VLKT~\cite{Liao_2022_CVPR}~\scriptsize({CVPR'22}) & 25.91 & 21.36 & 32.91 & 30.56 \\
		HOICLIP~\cite{ning2023hoiclip}~\scriptsize({CVPR'23}) & 29.47 & 25.53 & 28.47 & 32.99 \\
        ADA-CM~\cite{lei2023efficient}~\scriptsize({ICCV'23}) & 30.62 & 27.63 & 34.35 & 33.01 \\
        UniHOI~\cite{cao2024detecting}~\scriptsize({NeurIPS'23}) & 27.54 & 23.41 & 33.45 & 31.97 \\
        LogicHOI~\cite{li2024neural}~\scriptsize({NeurIPS'23}) & 29.79 & 25.97 & 34.93 & 33.17 \\
        CLIP4HOI~\cite{mao2024clip4hoi}~\scriptsize({NeurIPS'23}) & 31.59 & 28.47 & \textbf{35.48} & \underline{34.08} \\
        CMMP~\cite{lei2024exploring}~\scriptsize({ECCV'24}) & 31.07 & \underline{29.45} & 32.87 & 32.18 \\
        EZ-HOI~\cite{lei2024efficient}~\scriptsize({NeurIPS'24}) & \underline{31.38} & {29.02} & 34.15 & 33.13 \\
        \hline
    	\bm{$Ours$} (HOLa) & \textbf{32.69} & \textbf{30.61} & \underline{35.08} & \textbf{34.19} \\
		\bottomrule
	\end{tabular}
}
\caption{
	Quantitative comparison of zero-shot HOI detection with state-of-the-art methods in Rare First Unseen-Composition (RF-UC) setting on HICO-DET. HM denotes the harmonic mean. 
    }
    \label{tab: RF-UC}
\end{table}

\begin{table}[t]
\centering
\resizebox{1\columnwidth}{!}{
	\begin{tabular}{l | c c c c }
		\toprule
		{Method} & HM & Unseen & Seen & Full \\
		\hline
		GEN-VLKT~\cite{Liao_2022_CVPR}~(\scriptsize{CVPR'22}) & 15.42 & 10.51 & 28.92 & 25.63 \\
		HOICLIP~\cite{ning2023hoiclip}~(\scriptsize{CVPR'23}) & 21.28 & 16.20 & 30.99 & 28.53 \\
		UniHOI~\cite{cao2024detecting}~(\scriptsize{NeurIPS'23}) & 18.83 & 13.67 & 30.27 & 27.52 \\
        LogicHOI~\cite{li2024neural}~(\scriptsize{NeurIPS'23}) & 20.66 & 15.64 & 30.42 & 28.23 \\
        \hline
        \hline
        ADA-CM~\cite{lei2023efficient}~\scriptsize({ICCV'23}) & 32.40 & 33.26 & 31.59 & 31.87 \\
		CLIP4HOI~\cite{mao2024clip4hoi}~(\scriptsize{NeurIPS'23}) & 32.25 & 31.79 & {32.73} & {32.58} \\
        CMMP~\cite{lei2024exploring}~(\scriptsize{ECCV'24}) & 32.40 & \underline{33.76} & 31.15 & 31.59 \\
        EZ-HOI~\cite{lei2024efficient}~(\scriptsize{NeurIPS'24}) & \underline{32.66} & 33.28 & \underline{32.06} & \underline{32.27} \\
        \hline
    	\bm{$Ours$ } (HOLa)  & \textbf{34.65} & \textbf{36.45} & \textbf{33.02} & \textbf{33.59} \\
		\bottomrule
	\end{tabular}
}
\caption{
	Quantitative comparison of zero-shot HOI detection with state-of-the-art methods in Unseen-Object (UO) setting on HICO-DET. HM denotes the harmonic mean.}
\label{tab: UO}
\end{table}

\vspace{0.3cm}
\noindent \textbf{Unseen-Object Setting }
Table~\ref{tab: UO} presents the unseen-object (UO) setting performance comparison.
For a fair comparison, we focus on two-stage methods~\cite{lei2023efficient, mao2024clip4hoi, lei2024exploring, lei2024efficient}, all using the same object detector~\cite{carion2020end}. Our method outperforms all baselines across all metrics, achieving a 2.69 mAP gain on unseen classes.
This underscores our model's superior generalization to unseen object classes.

\begin{table}[t]
\centering
\setlength{\tabcolsep}{3pt} 
\footnotesize 
\resizebox{1\columnwidth}{!}{
	\begin{tabular}{c c c c c | c c c }
		\toprule
        \multirow{2}{*}[0.2ex]{\makecell[c]{Feat. \\ Dec.}} & \multirow{2}{*}[0.2ex]{\makecell[c]{Wgt. \\ Adapt.}}& \multirow{2}{*}[0.2ex]{\makecell[c]{LLM \\ Reg.}}&\multirow{2}{*}[0.2ex]{\makecell[c]{Txt. \\ Fusion}}&\multirow{2}{*}[0.2ex]{\makecell[c]{HO \\ Token}} & \multirow{2}{*}[-0.2ex]{\makecell[c]{Unseen}} & \multirow{2}{*}[-0.2ex]{\makecell[c]{Seen}}  & \multirow{2}{*}[-0.2ex]{\makecell[c]{Full}}  \\
        {}&{}&{}&{}&{}&{}&{}&{} \\
		\hline
        $\times$&$\times$&$\times$&$\times$&$\times$&23.58&31.55&30.43\\
		\textbf{$\checkmark$}&$\times$&$\times$&$\times$&$\times$&25.76&31.35&30.57\\
       \textbf{$\checkmark$}&\textbf{$\checkmark$}&$\times$&$\times$&$\times$& 25.47 & 33.59 & 32.46\\
		\textbf{$\checkmark$}&\textbf{$\checkmark$}&\textbf{$\checkmark$}&$\times$&$\times$&26.69&33.28&32.36 \\
		\textbf{\checkmark} &\textbf{$\checkmark$}&\textbf{\checkmark}&\textbf{\checkmark}&$\times$&27.19&33.55&32.66\\
        \textbf{$\checkmark$}&\textbf{$\checkmark$}&\textbf{\checkmark}&\textbf{\checkmark}&\textbf{\checkmark}&27.91&35.09&34.09   \\
		\bottomrule
	\end{tabular}
}
\caption{Ablation study for main components of our method in the unseen-verb zero-shot setting. ``Feat. Dec.'' denotes feature decomposition, ``Wgt. Adapt.'' denotes the weights adaptation, ``LLM Reg.'' denotes the LLM-derived action regularization, and  {``Txt. Fusion'' denotes the text fusion module.``HO Token'' denotes the human-object tokens in vision branch.}}
\label{tab: ablation_module}
\end{table}

\begin{figure}[t]
\centering
\includegraphics[width=0.47\textwidth]{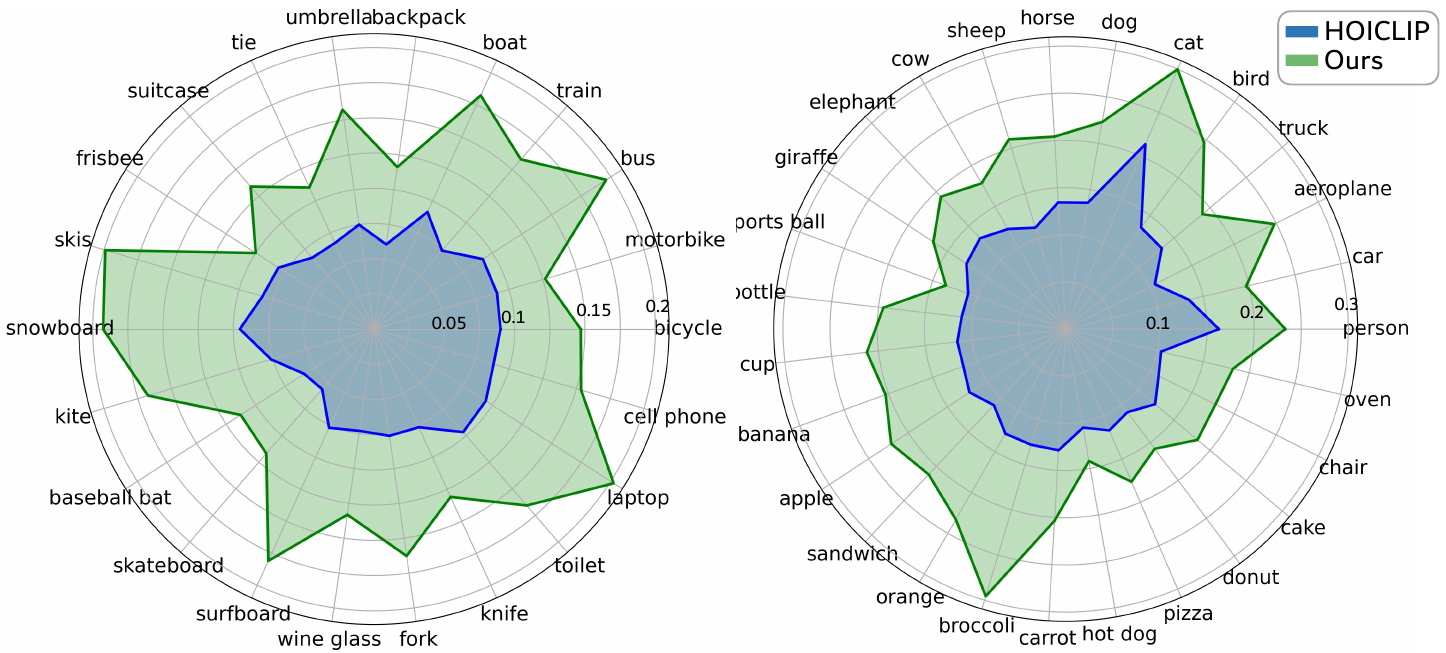}
\caption{ 
	Action dissimilarity comparison. The action dissimilarity ($AD$) of our reconstructed text features $ \bm{\hat{\mathrm{F}}} $ (green) is compared to that of the original VLM text features used by HOICLIP~\cite{ning2023hoiclip} (blue). Higher $AD$ values indicate better differentiation between actions involving the same object.
} \label{fig: act_dissimi}
\end{figure}

\subsection{Ablation Studies}
\noindent \textbf{Major Module Evaluation}
Table~\ref{tab: ablation_module} presents an ablation study on key modules under the unseen-verb zero-shot setting. The baseline (Row 1) excludes our proposed modules. 
To improve generalization, we introduce VLM feature decomposition and LLM-derived action regularization. VLM feature decomposition (Row 2) increases unseen mAP by 2.18 by reconstructing VLM features into basis features and weights, emphasizing class-shared information.
LLM-derived action regularization (Row 4) further improves unseen action distinction, adding a 1.22 mAP gain compared to weight adaptation without regularization (Row 3).

To refine action distinction, we introduce weight adaptation with LLM-derived action regularization and human-object tokens in the vision branch. The adaptable weights, optimized via action classification loss, improve seen-class performance by 2.24 mAP but slightly reduce unseen-class performance by 0.29 mAP due to reliance on seen-class ground truths. LLM-derived action regularization mitigates this, increasing unseen mAP by 1.22. Lastly, human-object tokens enhance interaction representation in the visual encoder, improving full mAP by 1.43.

To visualize enhanced action distinction, we compare our reconstructed HOI text features, derived from basis features and adapted weights, with the original VLM text features used by HOICLIP~\cite{ning2023hoiclip}.  
We define action dissimilarity for an object $o_{i_o}$ and its associated action set $\mathcal{A}_{i_o}$ as:  
$AD_{i_o} = \frac{1}{|\mathcal{A}_{i_o}|} \sum_{i, j \in \mathcal{A}_{i_o}, i \neq j}( 1 - \bm{{f}_{i}}^\top \bm{{f}_{j}})$
where $|\mathcal{A}_{i_o}|$ is the number of actions in $\mathcal{A}_{i_o}$, and $\bm{{f}_{i}}, \bm{{f}_{j}} \in \mathcal{R}^{d*1}$ are the text features of HOI classes ${\rm hoi}_i$ and ${\rm hoi}_j$.  
Fig.~\ref{fig: act_dissimi} plots $AD_{i_o}$ for different objects, showing that our reconstructed text features achieve higher action dissimilarity, leading to better differentiation between actions involving the same object.

\vspace{0.3cm}
\noindent \textbf{VLM Feature Decomposition and Adaptation } 
Table~\ref{tab: ablation_b_w} presents an ablation study on VLM feature decomposition and adaptation. To evaluate their effectiveness in isolation, we add only these two components to the baseline method (Row 3 in Table~\ref{tab: ablation_module}), excluding other proposed modules.
On the left, applying classification loss \( L_{\text{cls}} \) to both weights and basis features negatively affects both components during training. On the right, \( L_{\text{cls}} \) is applied only to weights, while \( L_{\text{fd}} \) is used for both ($\bm{{\mathrm{W}}}$, $\bm{\overline{\mathrm{B}}}$), leading to a 3.29 mAP gain on seen classes and 2.52 mAP on unseen classes over the baseline.  
These results show that updating both basis features and weights with \( L_{\text{cls}} \) reduces the effectiveness of class-shared information extraction, leading to a large performance drop.

\begin{table}[t]
\centering
\setlength{\tabcolsep}{3pt} 
\renewcommand{\arraystretch}{1.2} 
\footnotesize 
\resizebox{1\columnwidth}{!}{
	\begin{tabular}{c c | c c c !{\vrule width 1.2pt} c c | c c c  }
		\toprule
		{$\bm{\mathrm{W}}$}&{$\bm{\mathrm{B}}$}&Unseen & Seen & Full & {$\bm{\mathrm{W}}$}&{$\bm{\mathrm{B}}$}&Unseen & Seen & Full\\
		\hline
   $\bm{{\mathrm{W}}}$& $\bm{{\mathrm{B}}}$ & 22.95 & 30.30 & 29.27 & $\bm{{\mathrm{W}}}$ & $\bm{\overline{\mathrm{B}}}$ & \textbf{25.47} & \textbf{33.59} & \textbf{32.46}  \\
		\bottomrule
	\end{tabular}
}
\caption{Ablation study for weights and basis features optimization in the unseen-verb zero-shot setting. {$\bm{{\mathrm{X}}}$ means applying classification loss $L_{\rm cls}$ and feature decomposition loss $L_{\rm fd}$ in training to update $\bm{{\mathrm{X}}}$. $\bm{\overline{\mathrm{X}}}$ denotes applying only $L_{\rm fd}$. $\bm{\mathrm{X}} \in \{ \bm{\mathrm{W}}, \bm{\mathrm{B}} \}$.} 
}
\label{tab: ablation_b_w}
\end{table}

\vspace{0.3cm}
\noindent \textbf{LLM Description } 
We integrate LLM-generated descriptions for HOI and action classes in the language branch. Table~\ref{tab: ablation_llm} presents an ablation study on their impact.  
Without LLM descriptions, we use a fixed template: ``a person $\langle$ acting $\rangle$ a/an $\langle$ object $\rangle$'', the baseline unseen-class performance is 30.06 mAP (Row 1 in the Table~\ref{tab: ablation_llm}). 
Using our proposed method without the LLM (Row 2) improves unseen-class mAP by 4.60 over the baseline.
In contrast, adding LLM descriptions alone to the baseline provides only a 0.37 mAP boost, showing minimal standalone impact (Row 3).
With both our method and LLM descriptions (last row), unseen-class performance improves by 4.33 mAP compared to using LLM descriptions alone, demonstrating that our method’s effectiveness comes from its design rather than LLM reliance.  
Additional ablation studies for LLM description are provided in the supplementary.

\begin{table}[t]
\centering
\resizebox{0.85\columnwidth}{!}{
	\begin{tabular}{c c| c c c  }
		\toprule
		{Ours}&{LLM Description}&Unseen & Seen & Full \\
		\hline
  $\times$&$\times$ & 21.05&31.53&30.06 \\ 
  $\checkmark$&$\times$ & 25.65&33.86&32.71 \\
  \hline
        $\times$&$\checkmark$ &23.58&31.55&30.43\\ 
    $\checkmark$& $\checkmark$  & \textbf{27.91} & \textbf{35.09} & \textbf{34.09} \\
		\bottomrule
	\end{tabular}
}
\caption{Ablation study for LLM descriptions in the language branch.  ``Ours'' refers to our proposed HOLa modules. 
}
\label{tab: ablation_llm}
\end{table}

\begin{figure}[t]
\centering
\includegraphics[width=0.47\textwidth]{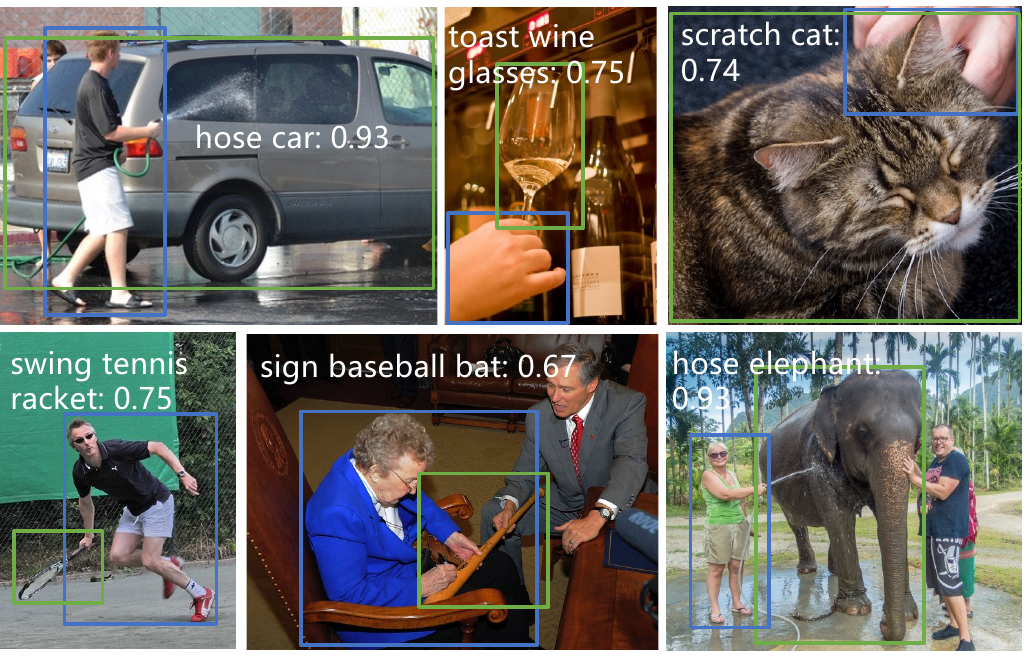}
\caption{ 
	Visualization of unseen HOI predictions in the unseen-verb setting on HICO-DET. 
} \label{fig: quali_main}
\end{figure}

\subsection{Qualitative Results}
Fig.~\ref{fig: quali_main} visualizes HOLa's predictions in the unseen-verb setting of HICO-DET. Our method successfully detects unseen actions like ``hose'', ``toast'' and ``swing'', showcasing strong generalization to unseen HOI classes.
This success stems from our low-rank decomposed feature adaptation, which captures class-shared information, while LLM-derived action regularization guides weight adaptation, for enhanced action distinction, especially for unseen actions.

\section{Conclusion}
We introduced HOLa, a novel Low-Rank Decomposed VLM Feature Adaptation method for zero-shot HOI detection, boosting generalization to unseen classes and action distinction.
HOLa employs low-rank factorization to decompose VLM text features into class-shared basis features and adaptable weights. The weights adapt each HOI class to improve action distinction, while the basis features, combined with weights, create a compact representation that preserves class-shared information, enhancing generalization.
To overcome the limitations of adapting weights solely from training data, we introduce LLM-derived action regularization to guide adaptation. Additionally, human-object tokens refine visual interaction representation, further improving action distinction.
HOLa achieves 27.91 mAP in the unseen-verb setting, establishing a new state-of-the-art in zero-shot HOI detection.

{
    \small
    \bibliographystyle{ieeenat_fullname}
    \bibliography{main}
}

\clearpage
\appendix
\section*{Supplementary Material}


\section{Vision Branch}
\label{sec:vision_branch}
\begin{figure*}[t]
	\centering
	\includegraphics[width=0.98\textwidth]{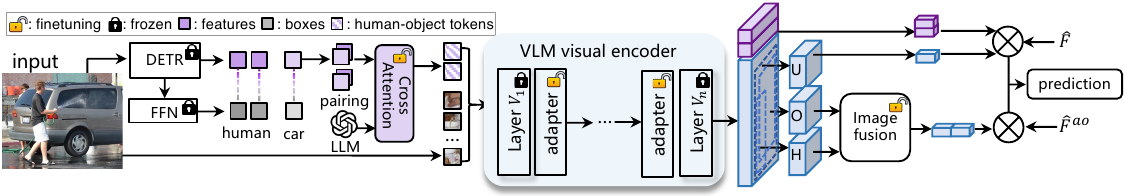}
	\caption{
		Overview of our vision branch. 
	} \label{fig: new_vis_enc}
\end{figure*}

The detailed vision branch is illustrated in Fig.~\ref{fig: new_vis_enc}. 
We first leverage a pre-trained DETR~\cite{carion2020end} model for object detection, identifying all detected humans and objects $h_i, o_j $, where $ i \in \{1, 2, \cdots, n_h\}, j \in \{1, 2, \cdots, n_o\}$. Here,  $n_h$  and  $n_o$ represent the total number of detected humans and objects, respectively.
For each detection, we extract human and object features from the DETR decoder, denoted as $f_{h_i}$ and $f_{o_j}$. We then generate all possible human-object feature pairs, represented as $(f_{h_i}, f_{o_j})$.
The human-object tokens are computed as \edit{}{:}
\begin{equation}
\label{eq: ho_token}
    T_{{\rm ho}_{ij}} = \frac{f_{h_i} + f_{o_j}}{2} + f_{{\rm ho}_{ij}}^{\rm spatial},
\end{equation}
where $f_{{\rm ho}_{ij}}^{\rm spatial}$  is derived from human-object bounding boxes, incorporating the center coordinates, width, height of each box, pairwise intersection-over-union (IoU), and relative area, which are then processed through an MLP together to obtain $f_{{\rm ho}_{ij}}^{\rm spatial}$.
Thus, $T_{{\rm ho}_{ij}}$ integrates both appearance and spatial cues to enhance interaction representation for each human-object feature pair $( f_{h_i}, f_{o_j})$. The complete set of human-object tokens is denoted as ${T}_{\rm ho}$, where ${T}_{\rm ho} = \{{T}_{{\rm ho}_{ij}} \mid 1 \leq i \leq n_h, 1\leq j \leq n_o  \}$.

To further incorporate interaction prior knowledge, we leverage an LLM to generate descriptions of human body configurations, object attributes, and their spatial relationships with humans. These descriptions are then encoded by the VLM text encoder to obtain prior knowledge features $f_{\rm ho}^{\rm pr}$.
An example of the generated descriptions used to capture human-object interaction prior knowledge is provided at the end of this section.

To integrate prior-knowledge features $f_{\rm ho}^{\rm pr}$ with human-object tokens, we design a cross-attention module:
First, down and up projection layers are used to reduce computational cost.
Next, human-object tokens serve as the query, while prior-knowledge features $f_{{\rm ho}}^{\rm pr}$ act as the key and value in the cross-attention mechanism.
Finally, a residual connection adds the input human-object tokens back to the cross-attention output, refining interaction representation while preserving the original information.

The output human-object tokens are concatenated with input image patches and fed into the VLM visual encoder, guiding it to focus on human-object interactions and improving action distinction. 
To enhance adaptability, we insert the adapter~\cite{lei2023efficient} between each layer of the visual encoder. 
The output includes adapted human-object tokens $\hat{T}_{\rm ho}$ and an image feature map $f_{\rm img}^{\rm glb} \in \mathcal{R}^{H \times W \times d}$. The final HOI image feature, denoted as $f_{\rm img} \in \mathcal{R}^{d \times 1}$ and used by the weight adaptation, is defined as follows:

\begin{equation}
    f_{\rm img} = \frac{1}{H*W} \sum_{i=1}^{H} \sum_{j=1}^W f_{\rm img}^{\rm glb}(i,j, :).
\end{equation}

We denote the detected bounding boxes for the human, object, and their union regions as $b_h$, $b_o$, and $b_u$, respectively.
To extract features focused on specific human-object interaction regions within the image, we first apply RoI pooling to obtain region-specific features of dimension $p \times p \times d$, where $p$ is set to 7.
We then apply spatial average pooling to each region-specific feature to obtain ${f}^h_{\rm img}, {f}^o_{\rm img}, {f}^u_{\rm img}$.

The image fusion module is designed to combine the human and object features ${f}_{\rm img}^h$ and ${f}_{\rm img}^o$, respectively. 
The image fusion process takes the $ {\rm concat}({f}_{\rm img}^h, \,{ f}_{\rm img}^o) $ as input and outputs $f^{\rm ho}_{\rm img}$. 
Here, ${\rm concat}$ denotes concatenation of two features along the first dimension. 
To reduce computational cost, the image fusion module incorporates down and up projection layers. 
The concatenated input features then pass through a self-attention module, integrating action and object visual features. 
Finally, a residual connection adds the input back to the output, refining the fusion while preserving input information.

According to Eq.(8) of the main paper, we compute the action prediction $s_a$. 
For convenience, we reproduce the equation below:
\begin{equation}
	\label{eq: score_act}
	\begin{aligned}
		s_a = \: & \gamma_1 * ( {\rm sim}({f}_{\rm img}^{u}, \bm{\hat{\mathrm{F}}} ) + {\rm sim}(\hat{T}_{\rm ho}, \bm{\hat{\mathrm{F}}} ) )* l_{u}^R + \\
		& \gamma_2 * {\rm sim}( f^{\rm ho}_{\rm img}, \bm{\hat{\mathrm{F}}^{\rm ao}}  ) * l_{ao}^R.  \\
	\end{aligned}
\end{equation}

Here is an example prompt used with an LLM to generate prior knowledge for the human-object pair $\langle$ human, car $\rangle$.

\emph{
Provide a detailed description of the physical relationship between a given human-object pair, focusing on various possible configurations and spatial relationships without assuming or naming specific interactions. For the pair (human, car), describing the following perspectives: 
}

\emph{
1. **Human Body Description:**  
   - Describe the positioning and orientation of key body parts (e.g., hands, feet, arms, legs, torso, head) in relation to the object.  
   - Highlight the possible roles of specific body parts (e.g., hands gripping, feet pressing, or knees bending) without specifying actions.  
}

\emph{
2. **Object Description:**  
   - Provide a clear and concise description of the object, focusing on its relative size, shape, and structure compared to the human in the image.  
   - Include details about key components (e.g., wheels, deck from skateboard) and their spatial relationship to the human body in various scenarios.  
   - Highlight how the object might be positioned (e.g., sliced, tilted, vertical) and how its components could interact with specific body parts of the human.  
}

\emph{
The goal is to provide a comprehensive pool of descriptive information for the human, object, and their possible configurations in various scenarios. Avoid limiting the scope by naming specific interactions or actions. Focus instead on a rich and versatile set of physical relationships. Focus on critical details, avoiding redundant or non-essential information to ensure clarity and precision.
}

The following is the prior knowledge generated by the LLM for the human–car pair: 

\emph{1. Human Body Description: (a) Positioning and Orientation:  -  Head : The human's head can be oriented in various directions relative to the car, such as facing towards the car, away from it, or to the sides. The head's angle might vary, possibly tilted up or down, depending on the human's position relative to the car's height, such as looking over the roof or under the chassis. -  Torso : The torso may be upright, leaning forward, or angled sideways. Its position could vary significantly depending on proximity to the car, such as standing next to it, bending over the hood, or leaning against the side. -  Arms and Hands : The arms might be extended towards the car or resting at the sides. Hands can be positioned near various parts of the car, such as the door handles, mirrors, or hood, suggesting a potential for contact or proximity. -  Legs and Feet : Legs could be positioned straight, bent, or spread apart, depending on the stance relative to the car. Feet might be placed firmly on the ground, perhaps transitioning from one side of the car to the other, or positioned close to the wheels or undercarriage. (b) Roles of Specific Body Parts:  -  Hands : The hands might appear poised to make contact with the car's surface, potentially hovering over or near tactile features like handles or mirrors. -  Feet : The feet might be aligned parallel to the car or angled towards it, suggesting readiness for movement or balance. -  Knees : Knees might be straight or slightly bent, indicating a stance that accommodates reaching or observing different parts of the car.}

\emph{2. Object Description: (a) Size, Shape, and Structure:  -  General Structure : The car is a relatively large object compared to a human, with a defined structure typically including a body, wheels, and windows. It is generally elongated with a front (hood), middle (cabin), and rear (trunk). -  Key Components :     -  Wheels : Typically four, located at the corners of the car, often near the ground and could be aligned with the human's feet or legs.    -  Body Panels : Including doors, hood, trunk, and roof, which vary in height and shape, potentially aligning with the human's torso or head.    -  Windows and Mirrors : Positioned at varying heights, likely matching the human's eye level or above, affecting the head's orientation. }

\section{Language Branch}
\label{sec:language_branch}
We leverage the LLM to generate descriptions for both HOI and action classes in the language branch.

\vspace{0.2cm}
\noindent \textbf{HOI Description Generation}
The descriptions for all given HOI classes are utilized to obtain the VLM text features $\bm{\mathrm{F}}$ through the VLM text encoder, as discussed in Section 3.1 of the main paper.
For a detailed illustration, consider the HOI class ``hosing a car'' as an example. 
The description generated for this HOI class is as follows:

\emph{``Hosing a car'' description: The photo shows a person using a hose to clean or wash a car. Common elements include water, soap, cleaning tools, and the person's attention to various parts of the car's exterior.
}

\noindent \textbf{Action Description Generation}
Additionally, we introduce action descriptions for all given action classes to obtain  action regularization, as discussed in Section 3.2 of the main paper.

Here is an example prompt provided to the LLM to generate a description for the action ``hose''.

\emph{
Please describe the visual features for action: hose.
The objects can be ``car, dog, potted plant, elephant''. Please summarize the common visual features for the action, shared by the objects. Use LESS than 250 characters for description. "
}

The following is the action description generated by the LLM for ``hose'':

\emph{The visual features for the action ``a person is hosing the object'' include: 1. Person holding a hose or water spray nozzle. 2. Object being hosed (e.g. car, dog, potted plant, elephant) in close proximity to the person. 3. Water spray or stream visible, potentially creating a mist or splash. 4. Person's arm or hand in motion, indicating the action of spraying water. 5. Background context suggesting an outdoor or yard setting.}

\section{Experiment Results}

\subsection{Implementation Details}

In weight adaptation, image and text fusion and human-object tokens design, all down projection layers reduce feature dimension from 512 to 64, while all up projection layers expands it back from 64 to 512. The head number of self attention and cross attention modules is 2.   The temperature of KL divergence used in our method is 0.1.
We use AdamW~\cite{loshchilov2017decoupled} as the optimizer and the initial learning rate is 1e-3.
For all experiments, our batch size is set as 64 on 4 A5000 GPUs.
Training takes 7 hours on 4 A5000 GPUs (22.5 GB VRAM each) with only 4.0M trainable parameters. Inference time is 82 ms per image.

We use three types of descriptions generated by foundation models: (1) HOI class descriptions from EZ-HOI~\cite{lei2024efficient}, generated using LLaVA~\cite{liu2023visual}. These descriptions are encoded by a VLM text encoder to produce $\bm{\mathrm{F}}$, as described in Section 3.1 (VLM Feature Decomposition and Adaptation) of the main paper. An example is also included in the HOI Description Generation subsection of the language branch (\cref{sec:language_branch}); (2) action descriptions for LLM-derived action regularization, generated using the LLaMA-3-8B model~\cite{dubey2024llama}, and used in the language branch (\cref{sec:language_branch}); and (3) Prior knowledge descriptions for human-object pairs, also generated by LLaMA-3-8B, and used in the vision branch (\cref{sec:vision_branch}).

\noindent 
\textbf{Datasets}
We evaluate our method on the HICO-DET dataset~\cite{chao2018learning}, a widely-used benchmark in human-object interaction detection. HICO-DET contains 47,776 images in total, consisting of 38,118 training images and 9,658 test images. The dataset includes 600 HOI classes combined from 117 action categories and 80 object categories.
We also provided the evaluation on the V-COCO~\cite{lin2014microsoft}, a subset of COCO, comprises 10,396 images, with 5,400 train-val images and 4,946 test images, and includes 24 action classes and 80 object classes. Note that V-COCO only contains evaluation under fully-supervised setting, but our focus is on the zero-shot HOI detection.

\subsection{Quantitative Results}

\begin{table}[t]
\begin{tabular}{l | c c c }
\toprule
		\multirow{2}{*}[-0.2ex]{Method}&\multicolumn{3}{c}{HICO-DET} \\
  \cline{2-4}
		{}&Full & Rare & Nonrare\\
  \hline
  One-stage Methods\\
  GEN-VLKT~(\scriptsize{CVPR'22})~\cite{Liao_2022_CVPR}&33.75&29.25&35.10\\
  EoID~(\scriptsize{AAAI'23})~\cite{wu2023end}& 31.11 &26.49& 32.49\\
  HOICLIP~(\scriptsize{CVPR'23})~\cite{ning2023hoiclip}&34.69&31.12&35.74\\
  LogicHOI~(\scriptsize{NeurIPS'23})~\cite{li2024neural} &\underline{35.47}&\underline{32.03}&\underline{36.22}\\
  UniHOI~(\scriptsize{NeurIPS'23})~\cite{cao2024detecting}&\textbf{35.92}&	\textbf{34.39}&	\textbf{36.26}	\\
  \hline
  \hline
  Two-stage Methods\\
  FCL~(\scriptsize{CVPR'21})~\cite{hou2021detecting}& 29.12 &23.67 &30.75\\
  ATL~(\scriptsize{CVPR'21})~\cite{hou2021affordance}& 23.81 &17.43& 25.72\\
    ADA-CM~\cite{lei2023efficient}~\scriptsize({ICCV'23})& 33.80& 31.72& 34.42\\
    CLIP4HOI~(\scriptsize{NeurIPS'23})~\cite{mao2024clip4hoi}&\underline{35.33}&\underline{33.95}&\textbf{35.75}\\
    CMMP~(\scriptsize{ECCV'24})~\cite{lei2024exploring}&33.24& 32.26& 33.53 \\
    \hline
  \bm{$Ours$ } (HOLa)  &\textbf{35.41}&\textbf{34.35}&\underline{35.73}\\
  \hline
  \hline
 {ADA-CM${}_{l}$}~\scriptsize({ICCV'23})  &38.40& 37.52& 38.66 \\
             {CMMP${}_{l}$}~\scriptsize({ECCV'24})  &38.14 &\underline{37.75} &38.25 \\
 {EZ-HOI${}_{l}$}~\scriptsize({NeurIPS'24}) & \underline{38.61}& 37.70& \underline{38.89}  \\
  \hline
  \bm{$Ours_l  $ }  (HOLa) &\textbf{39.05}&\textbf{38.66}&\textbf{39.17} \\
\bottomrule
\end{tabular}
\caption{Quantitative comparison of HOI detection with state-of-the-art methods in the fully-supervised setting on HICO-DET.  \bm{$Ours_l$ } denotes our scaled-up version utilizing the ViT-L/14
 backbone.}
     \label{tab: general_results}
\end{table}

\noindent \textbf{Fully Supervised Setting on HICO-DET}
We evaluate our method against HOI approaches with zero-shot HOI detection ability in the fully supervised settings , excluding methods that do not support unseen-action HOI detection.
Table~\ref{tab: general_results} demonstrates that our method sets a new state-of-the-art performance on the HICO-DET dataset in the fully supervised setting. 
Using the ViT-B backbone, the same as those used in existing methods~\cite{Liao_2022_CVPR, wu2023end, ning2023hoiclip, li2024neural, cao2024detecting, hou2021affordance, hou2021detecting, mao2024clip4hoi, lei2024exploring}, our method achieves a 35.41 mAP, surpassing all state-of-the-art two-stage HOI detection methods.
Switching to a ViT-L backbone further enhances performance, reaching 39.05 mAP. 
Although primarily designed to focus on zero-shot HOI detection and improve generalization to unseen classes, our method also shows competitive results in the fully supervised setting, underscoring its effectiveness across diverse evaluation scenarios.

\noindent \textbf{Fully Supervised Setting on V-COCO}
Our method also demonstrates competitive performance on the V-COCO dataset, achieving a 66.0 ${AP}^{S_2}_{role}$, achieving an improvement of 2.0 mAP over the current state-of-the-art method, CMMP~\cite{lei2024exploring}. Our ${AP}^{S_1}_{role}=60.3$.

\subsection{Ablation Study}

\noindent \textbf{Rank Selection for $\bm{\mathrm{B}}$ and $\bm{\mathrm{W}}$}
We conduct an ablation study on the rank \( m \) of the basis features and weights, as shown in Table~\ref{tab: ablation_rank}.
This study specifically explores the impact of the selected rank \( m \) on the performance, focusing solely on the feature decomposition module. 
Consequently, other components, such as the action prior and the action-object branch, were excluded from this analysis.

We initialize the weights and basis features using Principal Component Analysis (PCA). 
Specifically, we achieve reconstruction percentages of 0.80, 0.90, 0.95, and 0.98 for the original VLM text features, $\bm{\mathrm{F}}$. 
These percentages correspond to ranks of 17, 42, 71, and 119, respectively, in the obtained weights and basis features.

The evaluation results show that a rank 17 yields the highest unseen mAP (26.32), due to its compact representation that emphasizes class-shared features, enhancing generalization to unseen classes. 
However, this compactness leads to a drop in seen class performance, due to the loss of some detailed information from $\bm{\mathrm{F}}$.
Conversely, increasing the rank to 119 captures more class-specific details in the reconstructed features but diminishes the shared information across classes, leading to poorer unseen class performance. 
Consequently, we select the rank of 71 to optimally balance performance between seen and unseen classes.

\begin{table}[t]

\centering
	\begin{tabular}{c c | c c c }
		\toprule
		\multirow{2}{*}[-0.2ex]{\makecell[c]{reconstruction\\ score}} &\multirow{2}{*}[-0.2ex]{\makecell[c]{rank}}& \multicolumn{3}{c}{mAP}  \\
		\cline{3-5}
		{}&{}&Unseen & Seen & Full\\
		\hline
      0.80&17& 26.32&32.69& 31.80 \\
  0.90&42& 25.71&33.01& 31.98\\
  0.95&71& 25.47 & 33.59 & 32.46 \\
    0.98&119& 25.17&32.82& 31.75 \\
		\bottomrule
	\end{tabular}
\caption{Ablation study for the rank of basis features $\bm{\mathrm{B}}$ and weights $\bm{\mathrm{W}}$ in the unseen-verb zero-shot setting. }
\label{tab: ablation_rank}

\end{table}

\noindent \textbf{VLM Feature Decomposition Constraints}
We conducted an ablation study on the constraints for VLM feature decomposition as shown in Table~\ref{tab: ablation_fd}. 
The first row removes the orthogonal constraint $L_{\rm ort}$ on the basis features, leading to 1.10 mAP drop among unseen classes compared to the third row, indicating that the orthogonal constraint helps the basis features capture class-shared information more effectively, enhancing generalization to unseen classes.
Additionally, removing the sparsity constraint $L_{\rm sparse}$ (second row) lowers both seen and unseen performance, indicating that sparsity reduces  redundancy in the factorization, leading to a more compact representation.

\begin{table}[t]

\centering
	\begin{tabular}{ c c | c c c }
		\toprule
		 \multirow{2}{*}[-0.2ex]{\makecell[c]{$L_{\rm ort}$}}& \multirow{2}{*}[-0.2ex]{\makecell[c]{$L_{\rm sparse}$}}& \multicolumn{3}{c}{mAP}  \\
		\cline{3-5}
		{}&{}&Unseen & Seen & Full\\
		\hline
     $\times$&$\checkmark$& 26.81 & 34.70 & 33.60 \\
    $\checkmark$&$\times$& 27.47 & 34.45 & 33.48 \\
	$\checkmark$&$\checkmark$&27.91&35.09&34.09\\
		\bottomrule
	\end{tabular}
\caption{Ablation study for VLM feature decomposition constraints $L_{\rm ort}$ and $L_{\rm sparse}$ in the unseen-verb zero-shot setting. }
\label{tab: ablation_fd}
\end{table}

\noindent
\textbf{Semantic Loss}
We also design the semantic loss $L_{\rm sem}$ to preserve the distribution of pairwise cosine similarity among VLM text feature of each class.
The pairwise cosine similarity demonstrates the relationship between HOI classes indicated by VLM, which, trained on millions of data, generalizes these relationships to unseen classes. Unlike the original VLM features, which primarily emphasize object information and cluster different actions with the same object together, our method explicitly enhances action distinctions. To achieve this, we compute similarity only among HOI classes involving the same object, as shown in Eq.(\ref{eq: sem_loss_new}), with the mask M excluding interactions with different objects.

\begin{equation}
	\begin{aligned}
		& L_{\rm sem} = \:  D_{\text{KL}}[\: \frac{{\rm sim}( \bm{\hat{\mathrm{F}}} ,  \bm{\hat{\mathrm{F}}} )}{\tau}*M \: \parallel \: \frac{{\rm sim}( \bm{\mathrm{F}}, \bm{\mathrm{F}})}{\tau} *M\:] \\
		& + D_{\text{KL}}[ \frac{\: {\rm sim}( \bm{\hat{\mathrm{F}}^{\rm ao}} ,  \bm{\hat{\mathrm{F}}^{\rm ao}})}{\tau}*M \: \parallel \: \frac{{\rm sim}(\bm{\mathrm{F}}, \bm{\mathrm{F}})}{\tau}*M \:],
        \label{eq: sem_loss_new}   
	\end{aligned}
\end{equation}
where we apply a temperature coefficient $\tau$ in the KL divergence, setting $\tau = 0.1$ to emphasize action relationships that are underestimated in the original VLM features.
As shown in Table~\ref{tab: ablation_semloss}, without $L_{\rm sem}$, the overall performance in the unseen-verb setting decreases from 32.66 to 32.41 mAP, with a 0.48 mAP drop among unseen classes.

\begin{table}[t]
\centering
	\begin{tabular}{c | c c c }
		\toprule
		\multirow{2}{*}[-0.2ex]{\makecell[c]{$L_{\rm sem}$}}& \multicolumn{3}{c}{mAP}  \\
		\cline{2-4}
		{}&Unseen & Seen & Full\\
		\hline
        $\times$& 27.19&34.68& 33.63 \\
		$\checkmark$&27.91&35.09&34.09\\
		\bottomrule
	\end{tabular}
\caption{Ablation study for semantic loss in the unseen-verb zero-shot setting. }
\label{tab: ablation_semloss}
\end{table}

\noindent \textbf{Human-Object Tokens}
Table~\ref{tab: ablation_llm_hotoken} presents the ablation study on interaction prior knowledge generated by the LLM for human-object tokens $f_{\rm ho}$. In the first row, we remove this prior knowledge and replace cross-attention with self-attention process for $f_{\rm ho}$. The results indicate that interaction prior knowledge primarily improves seen-class performance. 
This is because the interaction prior knowledge provides all possible human body configurations, object attributes and their spatial relationships. During training, the model is guided by training data to select knowledge mainly for seen HOI classes. Consequently, this interaction prior knowledge does not obviously enhance unseen HOI performance. 

\begin{table}[t]

\centering
	\begin{tabular}{c | c c c   }
		\toprule
		\multirow{2}{*}[-0.2ex]{\makecell[c]{LLM-generated  \\ Prior Knowledge}}&\multicolumn{3}{c}{mAP}  \\
		\cline{2-4}
		{}&Unseen & Seen & Full \\
		\hline
     $\times$ & 27.90 & 34.58&33.65 \\
     $\checkmark$ & {27.91} & {35.09} & {34.09}  \\
		\bottomrule
	\end{tabular}
\caption{Ablation study for LLM description in human-object token design of the vision branch. ``None'' means no interaction prior knowledge generated from LLM. 
}
\label{tab: ablation_llm_hotoken}
\end{table}

Table~\ref{tab: ablation_hotoken} shows the ablation study on the components of human-object tokens $f_{{\rm ho}_{ij}}$. As defined in Eq.(\ref{eq: ho_token}), $f_{{\rm ho}_{ij}}$ consists of two components: human and object appearance features $\frac{f_{h_i} + f_{o_j}}{2} $ from DETR  and the spatial features $f_{{\rm ho}_{ij}}^{\rm spatial}$ from detected human and object bounding boxes. We found that we need to combine all components in human-object tokens for the best performance among both seen and unseen classes according to the results shown in Table~\ref{tab: ablation_hotoken}.

\begin{table}[t]

\centering
\renewcommand{\arraystretch}{1.2} 
	\begin{tabular}{c | c c c   }
		\toprule
		\multirow{2}{*}[-0.2ex]{\makecell[c]{$f_{{\rm ho}_{ij}}$}}&\multicolumn{3}{c}{mAP}  \\
		\cline{2-4}
		{}&Unseen & Seen & Full \\
		\hline
    $\frac{f_{h_i} + f_{o_j}}{2} $ & 27.37 & 34.42 & 33.43 \\
    $f_{{\rm ho}_{ij}}^{\rm spatial}$ & 27.59 & 34.63 & 33.64\\
    $\frac{f_{h_i} + f_{o_j}}{2} + f_{{\rm ho}_{ij}}^{\rm spatial}$ & {27.91} & {35.09} & {34.09}  \\
		\bottomrule
	\end{tabular}
\caption{Ablation study for human-object token design of the vision branch. ``None'' means no interaction prior knowledge generated from LLM. 
}
\label{tab: ablation_hotoken}
\end{table}

\noindent \textbf{Image Fusion}
Table~\ref{tab: ablation_imgfus} presents the ablation study for the image fusion module. Removing this module reduces performance from 34.09 to 33.10 mAP, highlighting its effectiveness. The image fusion module adapts and integrates separate action and object visual features, capturing more fine-grained information than human-object union region features. While  ${f}_{\rm img}^h$ ,  ${f}_{\rm img}^o$ , and  ${f}_{\rm img}^u $ share the same feature dimension,  ${f}_{\rm img}^h$  and  ${f}_{\rm img}^o$  focus on smaller, localized regions—human and object separately, rather than their combined union. This processing better preserves action and object details, ultimately improving performance.

\begin{table}[t]
\centering
	\begin{tabular}{c | c c c }
		\toprule
		\multirow{2}{*}[-0.2ex]{\makecell[c]{Image Fusion}}& \multicolumn{3}{c}{mAP}  \\
		\cline{2-4}
		{}&Unseen & Seen & Full\\
		\hline
        $\times$ & 26.46 & 34.19 & 33.10 \\
		$\checkmark$&27.91&35.09&34.09\\
		\bottomrule
	\end{tabular}
\caption{Ablation study for image fusion design in the unseen-verb zero-shot setting. }
\label{tab: ablation_imgfus}
\end{table}

\noindent \textbf{VLM Feature Decomposition and Adaptation}
Table \ref{tab: ablation_b_w} presents an ablation study on the VLM feature decomposition and adaptation.  {The first row serves as the baseline, where VLM feature decomposition is not applied, and no LLM-derived regularization are used. This baseline ensures that the ablation study specifically analyzes the impact of VLM feature decomposition.}
In the second row, we apply the only feature decomposition loss \( L_{\text{fd}} \) to update both weights and basis features ($\bm{\overline{\mathrm{W}}}$, $\bm{\overline{\mathrm{B}}}$). This improves unseen mAP by 2.18, indicating that feature decomposition enhances generalization to unseen classes.
Applying classification loss \( L_{\text{cls}} \) only to the basis features ($\bm{\overline{\mathrm{W}}}$, $\bm{{\mathrm{B}}}$), as in the third row, yields results similar to the second row. 
In the fourth row, adding classification loss \( L_{\text{cls}} \), supervised by ground truths from seen classes together with \( L_{\text{fd}} \) in the training process, to both weights and basis features results in performance degradation ($\bm{{\mathrm{W}}}$, $\bm{{\mathrm{B}}}$). 
This suggests that the updating of basis features from training data compromises essential class-shared information necessary for generalization, while the weights do not adapt effectively to distinguish actions within the HOI setting.
The last row shows the best results, where \( L_{\text{cls}} \) is applied only to the weights, while \( L_{\text{fd}} \) is used for both weights and basis features ($\bm{{\mathrm{W}}}$, $\bm{\overline{\mathrm{B}}}$). 
This configuration achieves balanced performance across seen and unseen classes, improving the seen mAP by 2.04 and the unseen mAP by 1.89 compared to the baseline.

\begin{table}[t]

\centering
	\begin{tabular}{c c | c c c }
		\toprule
		\multirow{2}{*}[-0.2ex]{\makecell[c]{$\bm{\mathrm{W}}$}}& \multirow{2}{*}[-0.2ex]{\makecell[c]{$\bm{\mathrm{B}}$}}& \multicolumn{3}{c }{mAP}  \\
		\cline{3-5}
		{}&{}&Unseen & Seen & Full \\
		\hline
 / & / & 23.58 & 31.55 & 30.43\\ 
 $\bm{\overline{\mathrm{W}}}$ & $\bm{\overline{\mathrm{B}}}$ & 25.76 & 31.35 & 30.57\\
  $\bm{\overline{\mathrm{W}}}$& $\bm{{\mathrm{B}}}$ & 25.84 & 31.19 & 30.44 \\
  $\bm{{\mathrm{W}}}$& $\bm{{\mathrm{B}}}$ & 22.95 & 30.30 & 29.27 \\
  $\bm{{\mathrm{W}}}$ & $\bm{\overline{\mathrm{B}}}$ & 25.47 & 33.59 & 32.46\\
		\bottomrule
	\end{tabular}
\caption{Ablation study for weights and basis features optimization in the unseen-verb zero-shot setting. {$\bm{{\mathrm{X}}}$ denotes applying classification loss $L_{\rm cls}$ and feature decomposition loss $L_{\rm fd}$ in training to update $\bm{{\mathrm{X}}}$. $\bm{\overline{\mathrm{X}}}$ denotes applying only $L_{\rm fd}$. $\bm{\mathrm{X}} \in \{ \bm{\mathrm{W}}, \bm{\mathrm{B}} \}$.} 
}
\label{tab: ablation_b_w}
\end{table}

\begin{figure}[t]
	\centering
	\includegraphics[width=0.48\textwidth]{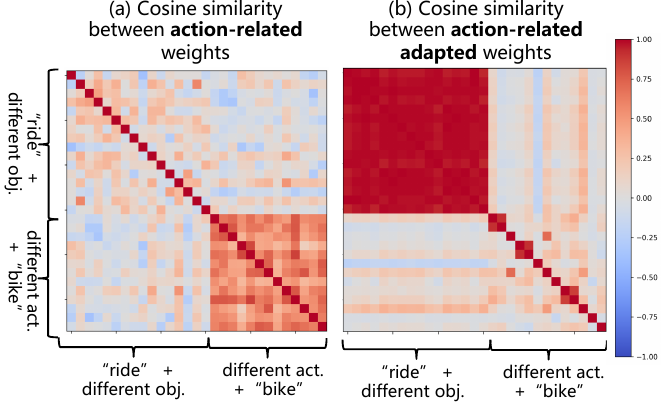}
	\caption{
		(a) Weight subset similarity visualization related to ``ride'' and ``bike'' HOI classes, \textbf{before weight adaptation}; (b) Adapted weight subset similarity visualization related to ``ride'' and ``bike'' HOI classes, \textbf{after weight adaptation}.  
	} \label{fig: ar_or_w_cmp_main}
\end{figure}

\noindent
\textbf{Weights for each training loss term}
Loss weights including $\alpha, \beta_1, \beta_2, \beta_3, \beta_4$ introduced in Section 3.4 of the main paper, are set to keep all loss terms on a comparable scale during early training, ensuring balanced contributions.  
Table~\ref{tab: ablation_hyploss} shows an ablation study where we vary one loss weight at a time while keeping the others fixed, where using comparable values for each loss term results in the best overall performance.

\begin{table}[ht]
\centering
\resizebox{0.9\columnwidth}{!}{
	\begin{tabular}{c c c c c| c c c }
		\toprule
		\multirow{2}{*}[-0.2ex]{\makecell[c]{$\alpha$}}& \multirow{2}{*}[-0.2ex]{\makecell[c]{$\beta_1$}}&\multirow{2}{*}[-0.2ex]{\makecell[c]{$\beta_2$}}&\multirow{2}{*}[-0.2ex]{\makecell[c]{$\beta_3$}}&\multirow{2}{*}[-0.2ex]{\makecell[c]{$\beta_4$}}& \multicolumn{3}{c }{mAP}  \\
		\cline{6-8}
		{}&{}&{}&{}&{}&Unseen & Seen & Full \\
		\hline
        \textcolor{blue}{320}&0.1&0.1&0.001&50& 27.50&33.95&33.05 \\ 
        80&\textcolor{blue}{0.5}&0.1&0.001&50&  28.32&34.35& 33.50  \\
        80&0.1&\textcolor{blue}{0.5}&0.001&50& 28.81& 34.64& 33.82 \\
        80&0.1&0.1&\textcolor{blue}{0.005}&50&  27.12&34.38& 33.36 \\
        80&0.1&0.1&0.001&\textcolor{blue}{250} & 27.87&33.69& 32.88 \\
        80&0.1&0.1&0.001&50&{27.91}&{35.09}&{34.09} \\
		\bottomrule
	\end{tabular}
}
\caption{Ablation study for training loss weights in the unseen-verb zero-shot setting. In each row, one loss weight is varied while others remain fixed. The changed value is shown in blue.
}
\label{tab: ablation_hyploss}
\end{table}

\noindent
\textbf{Visual Features for Human, Object and Union Regions}
We use three visual features from the image feature map $f_{\rm img}^{\rm glb}$ for HOI prediction: human ($f_{\rm img}^h$), object ($f_{\rm img}^o$), and union ($f_{\rm img}^u$) features. Ablation results in Table~\ref{tab: ablation_hou} show that using all three yields the best performance.

\begin{table}[ht]
\centering
\resizebox{0.65\columnwidth}{!}{
	\begin{tabular}{c| c c c }
		\toprule
		\multirow{2}{*}[-0.2ex]{}& \multicolumn{3}{c }{mAP}  \\
		\cline{2-4}
		{}&Unseen & Seen & Full \\
		\hline
        H+U&27.92&33.80&32.98 \\
        O+U& 27.58& 33.95& 33.06 \\
        H+O&27.36&34.81&33.76 \\
        H+O+U&{27.91}&{35.09}&{34.09} \\
		\bottomrule
	\end{tabular}
}
\caption{ Ablation study on visual features in the vision branch under the unseen-verb zero-shot setting. ``H'', ``O'', and ``U'' denote $f_{\rm img}^h$, $f_{\rm img}^o$, and $f_{\rm img}^u$, respectively.
}
\label{tab: ablation_hou}
\end{table}

\begin{figure*}[t]
\centering
\includegraphics[width=0.9\textwidth]{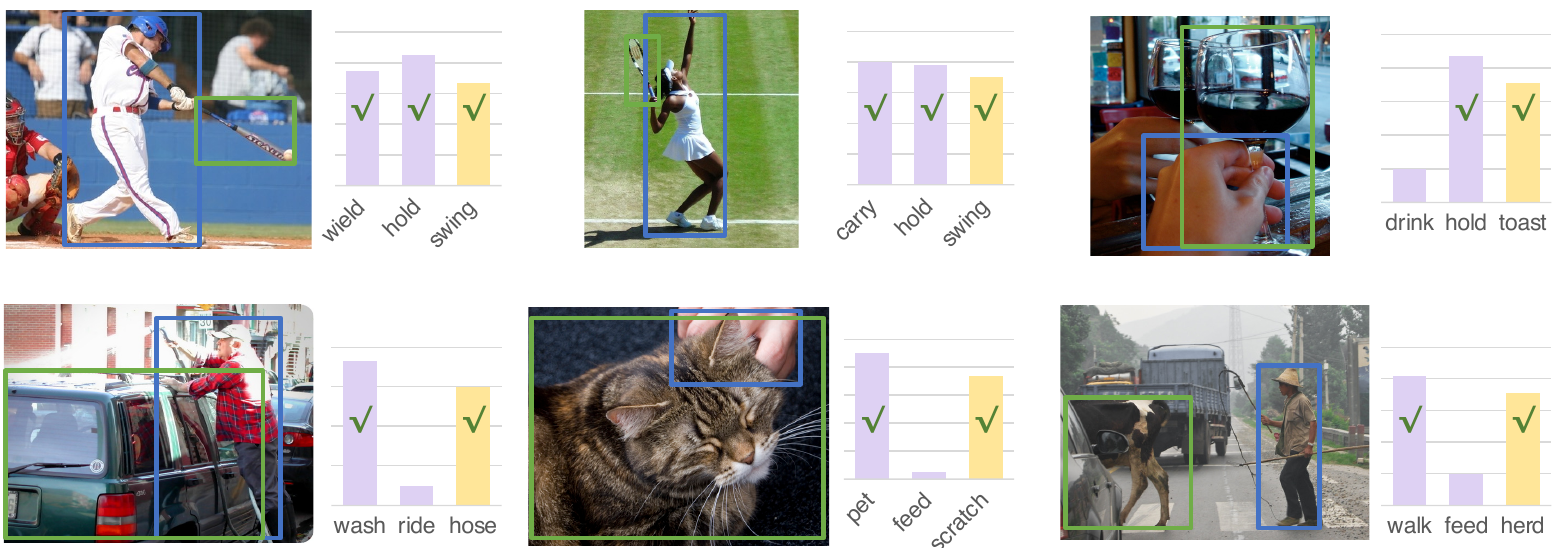}
\caption{ 
	Visualization of HOI predictions in the unseen-verb setting on HICO-DET. The purple bar indicates predictions for seen HOI classes and the yellow bar indicates predictions for unseen HOI classes.
} \label{fig: supp_uv_quali}
\end{figure*}

\noindent
\textbf{Weight Adaptation Visualization}
Here, we visualize and compare the weights $\bm{\mathrm{W}}$ before and after the weight adaptation process, especially focusing on the subset of $\bm{\mathrm{W}}$ applied with the LLM-derived action regularization, as discussed in the main paper Section 3.2.
The index set for the subset selection is defined as ${\mathcal{I}} = \{\,i \mid \bm{b}_{i}\in\mathbf{B}^{a}\,\}$, where $\bm{b}_{i}$ is the $i$-th row of the matrix $\mathbf{B}$ and also belongs to the subset $\mathbf{B}^{a}$
Before weight adaptation, the subset of $\bm{\mathrm{W}}$ is obtained by $ \{ {\bm{w}}_{i}' \mid i \in {\mathcal{I}} \}$, where ${\bm{w}}_{i}'$ is the $i$-th column of the matrix $\mathbf{W}$.
After weight adaptation, the subset is denoted as $ {\mathbf{W}}_{\mathrm{ar}} = \{ \hat{\bm{w}}_{i}' \mid i \in {\mathcal{I}} \}$, where $\hat{\bm{w}}_{i}'$ is the $i$-th column of the adapted matrix $\mathbf{W}$.

As shown in Fig.~\ref{fig: ar_or_w_cmp_main} (a), the subset before the weight adaptation contains limited action-specific information, as indicated by the low cosine similarities between weights for HOI classes associated with the action ``ride''. This suggests that shared information specific to the action ``ride'' is not well captured.
Moreover, the weights for classes involving the same object, ``bike'', show high similarity between each other, before weight adaptation. This demonstrates that in the original VLM feature space, actions linked to the same object tend to cluster together.
After our proposed weight adaptation, the weight subset $\bm{{\mathrm{W}}_{\rm ar}}$ show noticeably higher similarities among classes that share the ``ride'' action.

\subsection{Qualitative Results}

\begin{figure*}[t]
\centering
\includegraphics[width=0.9\textwidth]{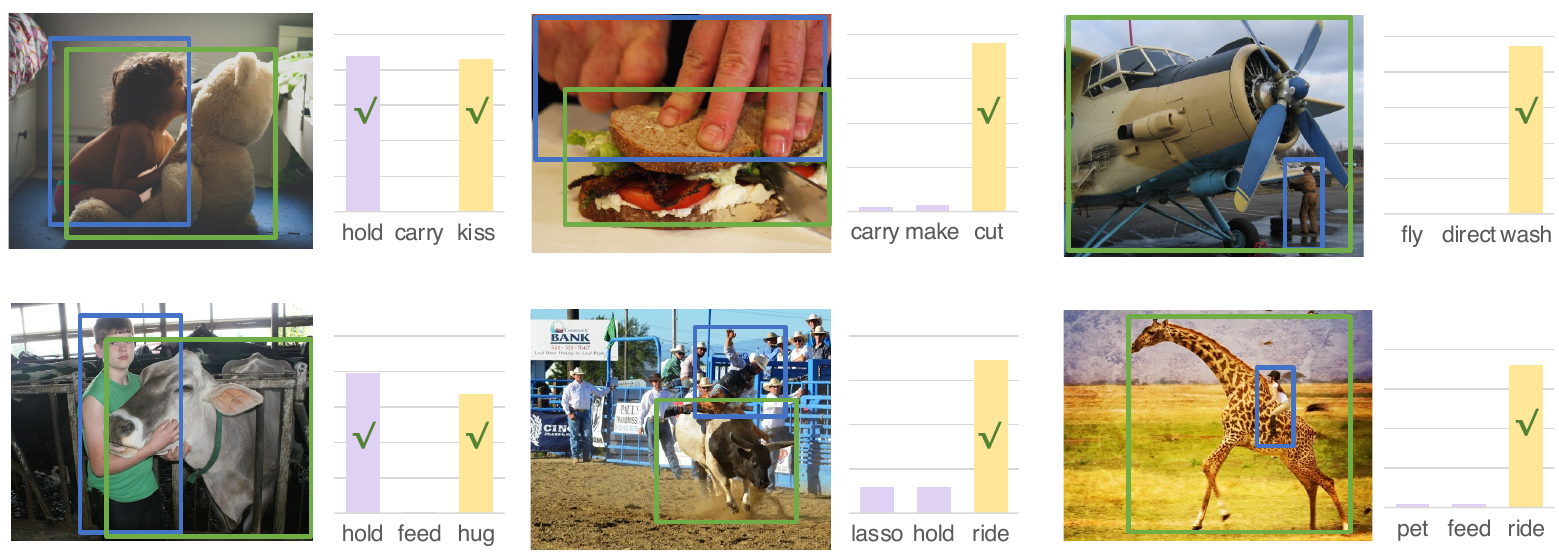}
\caption{ 
	Visualization of HOI predictions in the rare-first unseen-composition setting on HICO-DET. The purple bar indicates predictions for seen HOI classes and the yellow bar indicates predictions for unseen HOI classes.
} \label{fig: supp_rf_quali}
\end{figure*}

\begin{figure*}[!htbp]
\centering
\includegraphics[width=0.9\textwidth]{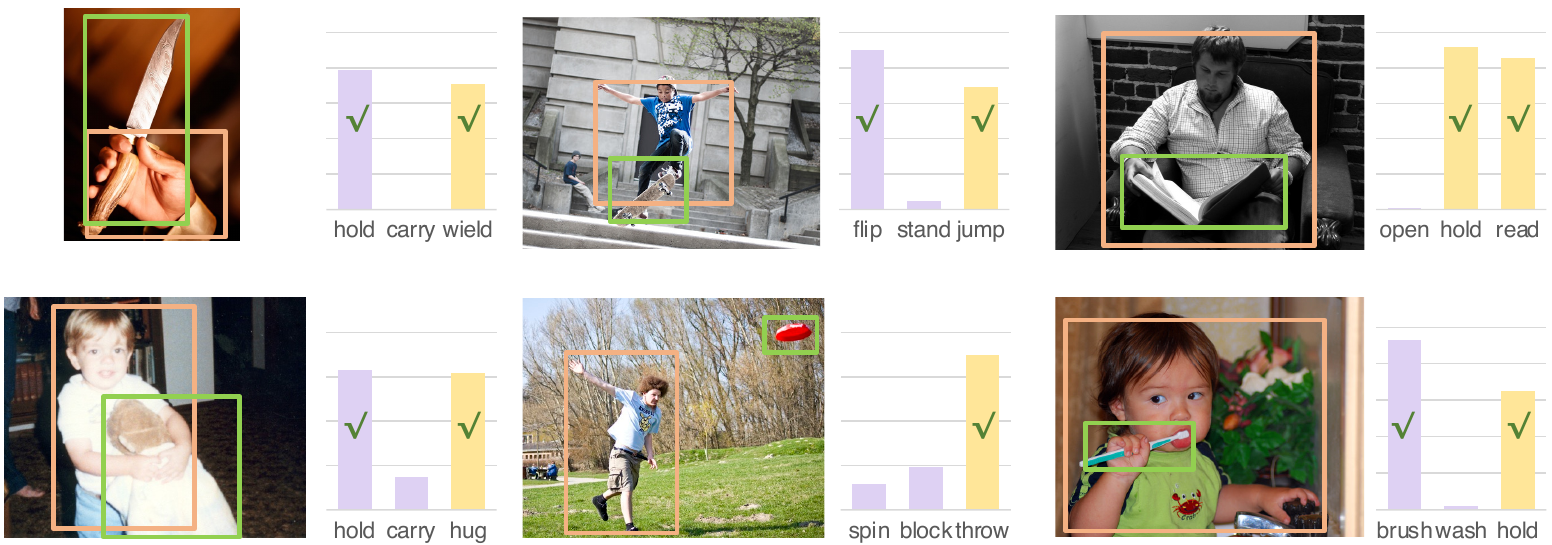}
\caption{ 
	Visualization of HOI predictions in the non-rare-first unseen-composition setting on HICO-DET. The purple bar indicates predictions for seen HOI classes and the yellow bar indicates predictions for unseen HOI classes.
} \label{fig: supp_nrf_quali}
\end{figure*}

\begin{figure*}[!htbp]
\centering
\includegraphics[width=0.9\textwidth]{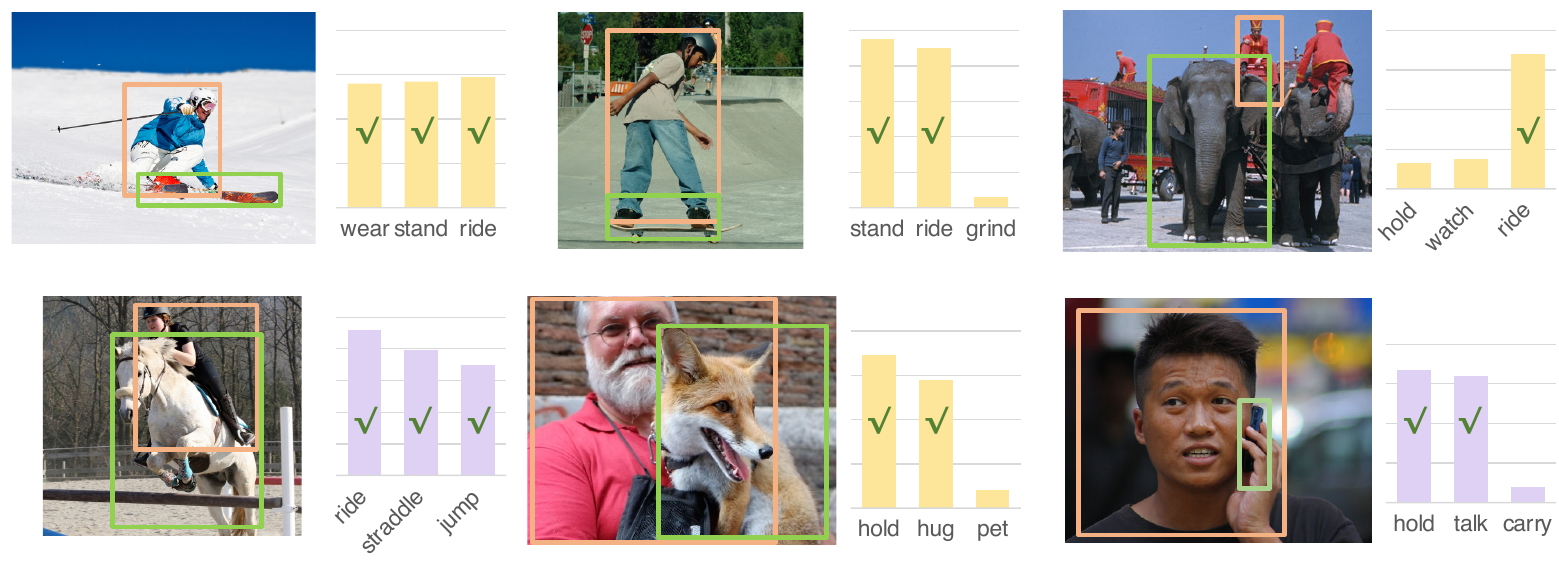}
\caption{ 
	Visualization of HOI predictions in the unseen-object setting on HICO-DET. The purple bar indicates predictions for seen HOI classes and the yellow bar indicates predictions for unseen HOI classes.
} \label{fig: supp_uo_quali}
\end{figure*}

We visualize our method's predictions across four settings in zero-shot HOI detection of HICO-DET: the unseen-verb setting in Fig.~\ref{fig: supp_uv_quali}, the rare-first unseen-composition setting in Fig.~\ref{fig: supp_rf_quali}, the non-rare-first unseen-composition setting in Fig.~\ref{fig: supp_nrf_quali} and the unseen-object setting in Fig.~\ref{fig: supp_uo_quali}. 
Our HOLa successfully identifies unseen HOI classes in various scenarios, demonstrating its generalization ability to unseen HOI classes.
This performance is due to our low-rank decomposed feature adaptation that emphasizes class-shared information, thereby enhancing generalization to unseen classes. Additionally, the incorporation of action priors helps reduce overfitting to seen classes.

\subsection{Controlability}
While the learned basis features in our low-rank decomposition is not directly interpretable, our method enhances controllability by restricting adaptation to a low-dimensional subspace, spanned by basis vectors $\mathbf{b}_{i} \in \bm{\mathrm{B}}$. In this subspace, explicit structures (e.g., orthogonality) are enforced and inspected, instead of modifying the features in the full VLM space.

\subsection{Limitations}
While our method achieves strong performance in zero-shot HOI detection, it relies on predefined unseen HOI class names, a standard requirement in zero-shot protocols~\cite{hou2020visual, hou2021affordance, hou2021detecting, wu2023end, ning2023hoiclip}. However, this dependency limits flexibility and scalability in real-world scenarios where such predefined classes may be unavailable.
To address this, our future work will focus on extending our approach to open-vocabulary HOI detection~\cite{Lei_2024_CVPR, yang2024open}.

\subsection{Future Work Exploration}
In our method, adaptation with low-rank decomposition is applied to the language branch, specifically on action and interaction features, to enhance generalization to unseen classes. This design leverages the availability of unseen class text descriptions during training, enabling the model to incorporate class-shared knowledge from both seen and unseen HOI classes.

Similar techniques could potentially be extended to object features in the language branch or to visual features. However, in standard two-stage HOI methods~\cite{hou2020visual,hou2021affordance,lei2023efficient}, object detection is typically handled by an off-the-shelf detector. As a result, the primary challenge in HOI detection lies in modeling unseen actions or novel action-object pairs, rather than object categories, where object generalization is addressed separately in open-vocabulary object detection. However, applying low-rank decomposition to object features may offer a promising direction to benefit open-vocabulary object detection as well.

Furthermore, visual features from unseen classes are not accessible under the standard zero-shot setting, making it infeasible to inject unseen information into the vision branch during training. Exploring decomposition strategies in the vision branch under settings with full or partial visual supervision is another promising avenue for future work.

\end{document}